\documentclass[twoside,11pt]{article}

\usepackage{amsmath,amsthm,amssymb,eurosym}
\usepackage{enumerate}
\usepackage{graphicx}
\usepackage{fancyhdr}
\usepackage{lastpage}
\usepackage{titlesec}
\usepackage[OT2,T1]{fontenc} 
\usepackage[russian,american]{babel} 
\usepackage{float}
\usepackage{bbm}
\usepackage[usenames,dvipsnames]{xcolor}
\usepackage{color}
\usepackage{soul} 
\usepackage{wrapfig}
\usepackage{verbatim}
\usepackage{algorithm}
\usepackage{algorithmic}
\usepackage{framed} 
\usepackage[utf8]{inputenc}
\usepackage[T1]{fontenc}
\usepackage{csquotes} 
\usepackage{url}
\usepackage{chngcntr}

\newcommand{\argmin}{\operatornamewithlimits{argmin}}

\theoremstyle{plain}
\newtheorem{theorem}{Theorem}
\newtheorem{corollary}{Corollary}
\newtheorem{conjecture}{Conjecture}
\newtheorem{lemma}{Lemma}
\newtheorem{proposition}{Proposition}

\theoremstyle{definition}
\newtheorem{definition}{Definition}

\usepackage{amsthm}
\let \proof \relax
\let \endproof \relax

\usepackage{gensymb}
\usepackage{amsmath, setspace}
\usepackage{enumerate}
\usepackage{hyperref}
\usepackage{cleveref}
\usepackage{natbib}
\usepackage[letterpaper,margin=1truein]{geometry}

\newtheorem{observation}{Observation}

\newtheorem*{restatethm}{Theorem}
\newtheorem*{restateprop}{Proposition}

\newcommand{\R}{\mathbb{R}}

\newcommand{\1}{\mathbf{1}}

\newcommand{\ra}{\rightarrow}

\renewcommand{\1}{\mathbf{1}}

\newcommand{\Li}{\mathcal{L}}

\usepackage{color}

\newcommand{\xhdr}[1]{{\bf #1.}}

\usepackage{chngcntr}
\usepackage{apptools}
\AtAppendix{\counterwithin{lemma}{section}}
\AtAppendix{\counterwithin{proposition}{section}}
\AtAppendix{\counterwithin{corollary}{section}}
\AtAppendix{\counterwithin{theorem}{section}}
\AtAppendix{\counterwithin{definition}{section}}

\newenvironment{proof}{\noindent {\bfseries Proof. }}{\qed}

\begin{document}

\title{Choosing to Rank}
  
\author{
  Stephen Ragain, Johan Ugander\\
  Management Science and Engineering\\
  Stanford University\\
  \texttt{sragain@stanford.edu}, \texttt{jugander@stanford.edu} \\
}

\maketitle

\begin{abstract}

Ranking data arises in a wide variety of application areas but remains difficult to model, learn from, and predict. Datasets often exhibit multimodality, intransitivity, or incomplete rankings---particularly when generated by humans---yet popular probabilistic models are often too rigid to capture such complexities. In this work we leverage recent progress on similar challenges in discrete choice modeling to form flexible and tractable choice-based models for ranking data. We study {\it choice representations}, maps from rankings (complete or top-$k$) to collections of choices, as a way of forming ranking models from choice models. We focus on the {\it repeated selection} ($RS$) choice representation, first used to form the Plackett-Luce ranking model from the conditional multinomial logit choice model. We fully characterize, for a prime number of alternatives, the choice representations that admit ranking distributions with unit normalization, a desirably property that greatly simplifies maximum likelihood estimation. We further show that only specific minor variations on repeated selection exhibit this property. Our choice-based ranking models provide higher out-of-sample likelihood when compared to Plackett-Luce and Mallows models on a broad collection of ranking tasks including food preferences, ranked-choice elections, car racing, and search engine relevance tasks.

\end{abstract}


\section{Introduction}

Ranking data arise in a wide variety of domains including recommender systems, competition, and social choice. Working with ranking data presents considerable computational challenges that stem from the complex structure of the space of permutations $S_n$ \citep{diaconis1988group}, also known as the symmetric group. Models of ranking data are essentially parametric families of distributions in this space, and many adopted models \citep{mallows1957non,plackett1968random} are highly simplistic out of computational necessity. Recent exploration into transforming rankings into pairwise comparisons \citep{khetan2016data}, a special type of choice data, has yielded insights into more tractable inference for existing ranking models. In this work, we consider generalized transformations of rankings into choice data, not just pairwise comparisons, and use these transformations to create novel, tractable choice-based models of ranking data. 

\xhdr{Discrete choice modeling}
In discrete choice modeling, we aim to model the behavior of an agent offered a {\it choice set} $S$ of alternatives, where $S$ is some subset of the universe $U$ of alternatives.
Discrete choice models model the selection probability for every possible choice set simultaneously, so a single instance of a discrete choice model $p$ defines a distribution $p(\cdot,S)$ over every $S \subseteq U$. For alternative $i \in S$, $p(i,S)$ gives the selection probability of alternative $i$ from choice set $S$. For example, a multinomial logit (MNL) choice model with parameters $\gamma_i > 0$ for every $i \in U$ has selection probabilities $p(i,S) \propto \gamma_i$ for every $i \in S$, and every $S \subseteq U$ \citep{luce1959individual}. 
Although the space of subsets of $[n]$ and the space of rankings of $[n]$ differ, discrete choice models and ranking models face similar combinatorial challenges that makes balancing flexibility and tractability a shared concern.

In this work, we conceptualize rankings as arising from a sequence of choices, allowing us to create novel ranking models from novel choice models. Furthermore, estimation of these novel ranking models can be done by simply estimating choice models on choice data implied by ranking data, making model inference tractable whenever the underlying choice model inference is tractable. Our approach produces models that significantly outperform standard models in out-of-sample predictions for a wide variety of empirical datasets including food preferences, search engine relevance rankings, automobile races, and ranked choice elections.

\xhdr{ Relation to Plackett-Luce and Mallows} 
Two seminal ranking models, the Plackett-Luce model and the Mallows model, can both be derived from choice models by thinking of rankings as built ``top-down" from choices, where each item in the ranking is viewed as being chosen from the set of alternatives ranked lower \citep{critchlow1991probability}. We term this generic process {\it repeated selection}. More concretely, for a ranking $\sigma \in S_n$, $\sigma^{-1}(1)$ is the top-ranked item according to $\sigma$, $\sigma^{-1}(2)$ is the item ranked second, and so forth, with $\sigma^{-1}(n)$ being the item ranked last. A repeated selection model would give the probability of $\sigma$ as the probability of making $n-1$ independent choices. First, $\sigma^{-1}(1)$ is chosen from all the alternatives. Next, $\sigma^{-1}(2)$ is chosen from all alternatives except $\sigma^{-1}(1)$. This process repeats until $\sigma^{-1}(n-1)$ is chosen from $\{\sigma^{-1}(n-1),\sigma^{-1}(n)\}$, completing the ranking. A choice model $p$ can assign probabilities to each of these choices. As each choice is made independently, the probability of $\sigma$ arising under repeated selection is simply the product of the choice probabilities under $p$.

Using different parametric choice models in conjunction with repeated selection, we obtain different ranking models. For example, the Multinomial Logit choice model (MNL) with repeated selection yields the Plackett-Luce ranking model \citep{plackett1968random,luce1977choice}. The Mallows ranking model is recovered when using repeated selection in conjunction with the following (reverse-engineered) choice model: the selection probability of alternative $i$ is geometric in the number of other alternatives $j$ in the choice set $S$ ranked above $i$ according to some reference ranking $\sigma_0 \in S_n$ \citep{qin2010new,critchlow1991probability}. 

In this work we consider applying repeated selection to other, richer choice models. Two recent choice models designed to better model context effects are the Pairwise Choice Markov Chain (PCMC)  \citep{ragain2016pairwise} and Context Dependent Random Utility Model (CDM) \citep{cdm}. By breaking a ranking down into a series of repeated choices and applying these richer choice models, we obtain ranking models that are flexible, tractable, straightforward to estimate. Importantly, we find that they are superior to Plackett-Luce and Mallows models in a wide variety of application areas. By decomposing the probabilities of rankings into the probabilities affiliated with each entry, we find that while our novel models perform similarly to Plackett-Luce on predicting the top entry of a ranking (i.e. $\sigma^{-1}(1)$), they are much better to predict subsequent top entries.

\xhdr{ Choice Representations}
While building a ranking front-to-back through repeated selection is perhaps the most intuitive way to model rankings as arising from choices, another natural model, repeated elimination, would instead build rankings back-to-front. These two approaches, repeated selection and repeated elimination, are far from exhaustive. As part of this work we introduce the general concept of {\it choice representations}, maps of rankings to collections of choices.

Given a choice representation $c$ and a choice model $p$, we define distributions on $S_n$ where the probability of a ranking $\sigma \in S_n$ is proportional to the product of the choice probabilities under $p$ of the choices output by $c(\sigma)$, i.e.,
\begin{equation}
P_{c,p}(\sigma) = \frac{1}{Z(c,p)} \prod_{(i,S) \in c(\sigma)} p(i,S),
\end{equation}
where $Z(c,p)$ is a normalization constant that can generally vary for different $c$ and $p$:
\begin{equation}
Z(c,p) = \sum_{\sigma \in S_n} \prod_{(i,S) \in c(\sigma)} p(i,S). 
\end{equation}
This framework generalizes repeated selection and repeated elimination, both of which are examples of choice representations. For example, repeated selection, denoted $RS$, can be written as $RS(\sigma) = \{(\sigma^{-1}(i),\{\sigma^{-1}(j)\}_{j =i}^n)\}_{i=1}^{n-1}$, where the $i$-th choice in $RS(\sigma)$ denotes the $i$-th ranked item being chosen from everything ranked below it. 

Given a choice representation $c$ and choice model $p$ with parameters $\theta$, we model the ranking data by fitting the parameters of the choice model to the choice data outputed by~$c$. Suppose we have training data $\sigma_1,\dots,\sigma_m$ and a parametric family $\{p_\theta\}_{\theta \in \Theta}$ of choice models, then choice representation $c$ and parameters $\theta$, the log likelihood $\ell_c$ of our data under choice representation $c$ is:
\begin{equation}
\ell_c(\theta ; \sigma_1,\dots,\sigma_m) =  \left( \sum_{j=1}^m \sum_{(i,S) \in c(\sigma_j)} \log p_\theta(i,S) \right) -  m \log Z(c,p_\theta). 
\end{equation}

This expression clearly decomposes into the log likelihood of choice data under $p_\theta$ and the term $m \log Z(c,p_\theta)$, allowing us to maximize the log likelihood for our choice-based ranking model as long as we can both maximize the log likelihood of the choice data and efficiently compute $\log Z(c,p_\theta)$. 

The main theoretical contribution of our work (Theorem~\ref{thm:Z=1}) is to show, if $c$ is in a family of choice representations including both repeated selection and repeated elimination, then $Z(c,p)=1$ for any $p$. This result allows us to train these ranking models simply by training a choice model on the corresponding choice data.
Furthermore, we show (Theorem~\ref{thm:Z=1revprime}) that, under a natural assumption for our choice representations and for rankings of a prime number of items, {\it only if} $c$ can be written as composition of repeated selection with some permutation does $Z(c,p)=1$ for all choice models $p$. This result strongly motivates the study and use of repeated selection and repeated elimination, as they are the most natural choice representations in this family. We believe that the requirement for a prime number of alternatives $n$ is an artifact of our analysis, and conjecture that Theorem~\ref{thm:Z=1revprime} holds for all composite $n$ as well.

\xhdr{ Mixed length top-$k$ rankings}
An attractive practical benefit of our approach is that choice representations have natural extensions from full rankings down to partial rankings, which allows us to easily assign probabilities to top-$k$ rankings with our framework, where only the top $k$ of the $n$ alternatives in $U$ are ranked. We derive normalization results for top-$k$ rankings under both repeated selection and repeated elimination choice representations. These normalization results again allow us to estimate these top-$k$ ranking distributions by estimating choice models on choice data obtained from a training set of top-$k$ rankings. 

Many practical ranking datasets consist of mixed-length top-$k$ rankings. Because our framework easily extends to the top-$k$ case, it provides a principled and straightforward way to handle such empirical data. For example, in the ranked choice election data we study, voters were allowed to list as many of the candidates as they wanted to vote for, from most to least desirable. Our suite of results allows us to train a single choice model that simultaneously defines all top-$k$ distributions.

\xhdr{Empirical results}
We perform extensive empirical evaluations of our choice-based ranking models using real-world data, focusing both on predicting out-of-sample full and top-$k$ rankings as well as predicting sequential entries of rankings as the top entries are revealed. Our datasets span ranked choice voting from elections, lists of ranked sushi preferences, Nascar race results, and search engine results.  We find that the flexible choice-based ranking models we introduce in this work---based on the PCMC and CDM choice models---achieve significantly higher out-of-sample likelihood compared to the Plackett-Luce and Mallows models across a wide range of applications.  This collection of datasets demonstrates the broad efficacy of our approach across not only application areas but also differing dataset characteristics: these datasets range greatly in size, number of alternatives, how many rankings each alternative appears in, and uniformity of ranking length. 

\xhdr{Related Work}
Discrete choice theory has found wide applicability in domains that model consumer behavior, including demand forecasting \citep{mcfadden1986choice,ben1985discrete}, assortment optimization \citep{desir2016assortment}, and revenue management \citep{talluri2004revenue}. There is a rich history of formulating discrete choice models as arising from underlying preference rankings~\citep{farias2013nonparametric,blanchet2016markov,van2017expectation}. Meanwhile, we turn this connection around to build ranking models that arise from underlying choice preferences.

Our work on choice representations generalizes the seminal connection between the Plackett-Luce (PL) ranking model from the Multinomial Logit (MNL) choice model, whereby the PL model can be viewed as building rankings top-down though repeated applications of the MNL model \citep{luce1959individual,plackett1968random}. The Plackett-Luce model has also been studied under the monikers {\it rank-order logit} and {\it exploded logit} \citep{hausman1987specifying,allison1994logit}, where inference for complete, incomplete, and top-$k$ rankings as well as rankings with ties.  The notion of a choice representation fits within the framework of {\it generalized rank breaking} \citep{soufiani2013generalized,negahban2018learning,khetan2018generalized}, where the statistical and computational tradeoffs of fitting full ranking data versus some subset of the partial orders implied by the rankings are considered, e.g.~pairwise comparisons.

When applying repeated selection and repeated elimination to recent choice models in order to develop new ranking models, we specifically focus on two discrete choice models. The Pairwise Choice Markov Chain (PCMC) model frames choice probabilities as the stationary distribution a of continuous time Markov chain on the choice set \citep{ragain2016pairwise}, and is among a recent of wave in modeling work in discrete choice theory using Markov chains \citep{blanchet2016markov,negahban2016rank}. We also consider the Context Dependent Model (CDM), which models choices as arising from pairwise utilities within a choice set 
\citep{cdm}. Both the PCMC and CDM models include the MNL choice model as a special case, so repeated selection with these models can be conceived as two different generalizations of the Plackett-Luce ranking model. 

Axiomatic approaches are important in the development and study of discrete choice models. For example, the MNL model can be derived from Luce's choice axiom \citep{luce1959individual}, commonly referred to as the Independence of Irrelevant Alternatives (IIA). The flexibility of the PCMC and CDM choice models arise largely by eschewing many common choice axioms, but they still exhibit important structure through less rigid choice axioms such as {\it uniform expansion}~\citep{yellott1980generalized}. As part of this work we develop two weaker versions of IIA, {\it local IIA} and {\it nested IIA}, which we show hold for some PCMC and CDM models. We demonstrate tha repeated selection with choice models having these axioms yields ranking distributions with interesting theoretical structure. 

Another broad and important family of choice models known as Random Utility Models (RUMs) affiliate a ``utility'' distribution with each alternative and define the selection probability of an alternative as the probability that a draw from its utility distribution is the largest within the choice set. A natural ranking model arises from a RUM choice model by ordering the alternatives sorted by their utility, a process of {\it joint selection}. Joint selection transforms the MNL choice model into the Plackett-Luce models (the same correspondence as for repeated selection), but yields different ranking models for other RUMs \citep{block1960random}. Joint selection can only be applied to RUMs, and neither the CDM or PCMC choice model are RUMs \citep{ragain2016pairwise,cdm}. Some work has found success using the joint selection distribution of RUMs for modeling rankings \citep{azari2013preference}, but our focus in this work is on choice-based models of ranking.

A common approach to modeling ranking data, though not typically viewed as choice-theoretic, involves characterizing probability distributions over the space of rankings in terms of distances away from a reference permutation on the symmetric group \citep{diaconis1988group,marden1996analyzing}. The most popular such distance, Kendall's $\tau$ distance $\tau(\sigma,\sigma_0)$, counts the number of ``inversions'' between two rankings: the number of pairs $i<j$ where $\sigma(i)<\sigma(j)$ but $\sigma_0(i)>\sigma_0(j)$. Kendall's $\tau$ underlies the popular Mallows model of ranking, which assigns each permutation a probability that decreases exponentially in some scaling of its Kendall's $\tau$ distance from reference permutation $\sigma_0$ \citep{mallows1957non}.
While popular, 
learning a Mallows model from data poses many challenges. Computing the most likely $\sigma_0$ from some input ranking data $\sigma_1,\dots,\sigma_k$ is equivalent to Kemeny rank aggregation, which is known to be a NP-hard problem for $k\geq 4$ input rankings \citep{dwork2001rank}.

Distance-based ranking models can be derived as repeated selection from a choice model when the underlying distance has a certain decomposition \citep{critchlow1991probability}. The Mallows model can be derived as repeated selection by decomposing $\tau$ into the inversions caused by placing each subsequent alternative ahead of those ranked after it \citep{qin2010new}. This decomposition give a greedy approximation algorithm which produces a locally (under $\tau$) optimal reference permutation for top-$k$ lists.

Lastly, we note the large body of work on ``learning to rank'' in the information retrieval literature \citep{yue2007support,kenyon2007rank}, which focuses on outputting a ranking to maximize metrics such as machine average precision (MAP), discounted cumulative gain (DCG), or mean reciprocal rank (MRR). Many learning to rank methods highlight the use of features of ranked alternatives or rankers or external quality scores \citep{yue2007support,sculley2010combined}. Our present work differs from the typical learning to rank framework both because we are learning a distribution over rankings (rather than outputting a single ranking) and because we do not focus on using relevance labels.

\section{Choice representations}
\label{sec:rsre}

In this section we formalize our translation for models of discrete choice data into models of ranking data. We first define choice models, choice representations, and ranking distributions based on choice representations. We then define repeated selection and repeated elimination in Section~\ref{sec:rsresub}. Label-invariance is defined as a property of a choice representation in Section~\ref{sec:labelinvar}, which allows us to state and prove our main results about unit normalization. Pairwise choice representations and choice representations for top-$k$ lists are discussed in Sections~\ref{sec:pairltr} and~\ref{sec:partialltr}, respectively. Section~\ref{sec:learning} concludes by establishing the role that unit normalization plays in facilitating maximum likelihood estimation of choice-based ranking models. 

Throughout this work a {\it ranking} $\sigma$ of a universe of alternatives $U$ with $|U|=n$ items is a bijection from $U$ to $[n]$, where $[n]=\{1,\dots,n\}$. We will typically think of $U$ as indexed by $[n]$, so that the space of rankings is the symmetric group $S_n$, the set of bijections from $[n]$ to itself. We will use $\sigma \in S_n$ to represent a permutation or ranking and $\sigma^{-1}$ to represent the unique inverse mapping with $\sigma(\sigma^{-1}(k))=k$ for all $k \in [n]$. Intuitively, $\sigma^{-1}(k)$ is the item ranked at position $k$ by $\sigma$. A {\it choice} is an ordered pair $(i,S)$ where $i$ is an element chosen from subset $S$ of the universe of alternatives $U$. In this work, we will only consider choices where the choice set $S$ has at least two alternatives, i.e. $|S| \geq 2$. A {\it choice probability} $p(i,S)$ is the probability of choosing $i$ from $S$ such that $\sum_{i \in S} p(i,S)=1$. We write $p_\theta(i,S)$ when the choice model is parameterized by some parameter(s) $\theta$.

We will first transform a model for choice into a model for ranking. In doing so, we will use choice probabilities for every subset $S$ of the universe $U$. We define a choice model as referring to the collection of distributions over all subsets.

\begin{definition}
A {\bf choice model} $p$ is a collection of probability distributions $p(\cdot,S)$ over all subsets $S$ of some universe of alternatives $U$. 
\end{definition}

Our primary vehicle for transforming ranking models into choice models is to transform rankings into sets of choices. To turn a ranking into a set of choices, we'll use a choice representation, defined here.

\begin{definition}
 A {\bf choice representation} is a mapping $c$ from a ranking $\sigma$ of alternatives in $U$ to a non-empty set of choices $(i,S)$.
\end{definition}

We can think of a ranking $\sigma$ as resulting from a series of choices between alternatives in $U$, with the choice representation $c$ giving us the translation from rankings into corresponding choices. Several choice representations that map to pairwise choices (only) arise in the literatures on ranking from pairwise comparisons~\citep{rajkumar2014statistical,agarwal2016ranking} and  ``rank breaking'' \citep{soufiani2013generalized,negahban2018learning}, which considers, e.g.,~the set of pairwise choices implied by $\sigma$ as $PW(\sigma) = \{(i,\{i,j\}) : \sigma(i) <\sigma(j) \}$. For example, if $\sigma$ ranks the set $\{1,2,3\}$ and $\sigma(1) = 2,\sigma(2)=1,\sigma(3) = 3$, this choice representation gives $PW(\sigma) = \{(2,\{1,2\}),(2,\{2,3\}), (1,\{1,3\})\}$.
 
The work on rank breaking has leveraged these representation of rankings as collections of pairwise choices to develop generalized method of moments estimators for model parameters \citep{hajek2014minimax,soufiani2013generalized,chen2015spectral}. One contribution of our work is to generalize rank breaking to choice representations other than just pairwise choices, thereby opening up a broader range of connections between ranking and choice. The pairwise choice representation $PW$ is discussed further in Section~\ref{sec:pairltr}.

Our use of general choice sets allows us to naturally extend the rank breaking concept to apply it to models that violate the independence of irrelevant alternatives where choices from larger sets can not be reduced to pairwise comparisons. The heart of our ``choosing to rank'' approach is thus the following translation from a choice model and a choice representation to a ranking distribution.

\begin{definition}
\label{def:cp}
For a choice representation $c$ and discrete choice model $p$ over $n$ alternatives, the {\bf $c_p$ distribution} over rankings has probabilities
$$
P_{c,p}(\sigma)  = \frac{1}{Z(p,c)} \prod_{(i,S) \in c(\sigma)} p(i,S),
$$ 
which defines a ranking distribution up to a {\it normalization constant} $Z(p,c)$: 
\begin{equation}
\label{eq:z}
Z(p,c) = \sum_{\sigma \in S_n} \prod_{(i,S) \in c(\sigma)} p(i,S).
\end{equation}
\label{def:c-p}
\end{definition}
\noindent When the choice model comes from a clear parametric family $p_\theta$ for $\theta \in \Theta$, we bypass $p$ in the notation: $P_{c,\theta}(\sigma)  \propto \prod_{(i,S) \in c(\sigma)} p_\theta(i,S)$. 

\subsection{Repeated selection and repeated elimination}
\label{sec:rsresub}
We proceed to define repeated selection ($RS$) as perhaps the most intuitive choice representation. For a choice model $p$, $RS$ envisions $\sigma^{-1}(1)$ as selected according to $p(\cdot,U)$, $\sigma^{-1}(2)$  according to $p(\cdot,U - \sigma^{-1}(1))$, and so forth. We also define the natural complement of $RS$, repeated elimination ($RE$).

\begin{definition} 
{\bf Repeated selection} (RS) is the choice representation 
$$
{RS}(\sigma) = \{(\sigma^{-1}(i),\{\sigma^{-1}(i),\sigma^{-1}(i+1),\dots,\sigma^{-1}(n)\})\}_{i=1}^{n-1}.
$$
\end{definition}

\begin{definition}
\label{def:re}
{\bf Repeated elimination} (RE) is the choice representation 
$$
RE(\sigma) = \{(\sigma^{-1}(i),\{\sigma^{-1}(i),\sigma^{-1}(i-1),\dots,\sigma^{-1}(1)\})\}_{i=2}^{n}.
$$
\end{definition}

A choice representation $c$, which maps ranking data to choice data, can also be thought of as implicitly mapping a choice model to a ranking model. Choice models typically belong to a (parametric) family of models $\mathcal M = \{p_\theta\}_{\theta \in \Theta}$. As such, we use the notation $c_{\mathcal M}$ to refer to a family of $c_p$ distributions (where a $c_p$ distribution is defined above in Definition~\ref{def:cp}). For example, for the repeated selection choice representation $RS$ and the multinomial logit (MNL) family of models we write $RS_{MNL} = \{ RS_{p} :~p~\text{is a MNL choice model}\}$. We use the phrase ``repeated selection with MNL" to refer to $RS_{MNL}$. 

Repeated selection was first proposed by Luce as the ``ranking postulate,'' accompanied by a theorem establishing that when Luce's choice axiom holds for the choice model, the probability that $x \in U$ is ranked above $y \in U$ under the ranking model is the same as the probability $x$ is selected from $\{x,y\}$ \citep{luce1959individual}.  

The past work most in the spirit of repeated elimination is the seminal Elimination by Aspects (EBA) choice model \citep{tversky1972elimination}, where it is notable that EBA is usually thought of primarily as a choice model and not a ranking model. In the EBA model, each alternative $i \in U$ has some set of ``aspects," and choices are made by randomly choosing an aspect shared by some of the alternatives and eliminating all of the alternatives lacking that aspect. 

In this work we focus on repeated selection ($RS$) and repeated elimination ($RE$) as the primary choice representations of interest. Not only are repeated selection and repeated elimination the most natural ways to translate rankings into choices, as evidenced by their connections to existing literature, but we will also show that they belong to a relatively small family of choice representations for which the normalization constant $Z(p,c)=1$ for all choice models $p$, which greatly simplifies many computational issues. The choice representations $RS$ and $RE$ are by far the two most natural representations in this family.

\subsection{Label-invariance}
\label{sec:labelinvar}

 In models of ranking, {\it label-invariance} is a natural property of models for rankings that simply allows the items ranked to be re-indexed without changing the model class~\citep{critchlow1991probability}. Equivalently, label-invariance means that our model will not be affected by how we choose to index the set of alternatives with $[n]$. Here we extend the notion of label-invariance to choice representations and show that label-invariant choice representations give a powerful, useful property to any $c_p$ distribution: unit normalization whereby $Z(p,c)=1$ for all $p$. We leverage unit normalization to efficiently maximize the likelihood of choice-based ranking models, given ranking data, in Section~\ref{sec:learning}. 
We also show that when ranking a prime number of items, the only label-invariant choice representations that have a normalization constant of 1 for all choice models are relabelings of repeated selection ($RS$). 

Informally, label-invariance is the idea that if we relabeled the items in $U$ and applied that relabeling to all of our rankings, the choices arising from the relabeled rankings should be the same as if we had simply applied the relabeling to each choice and choice set. More formally, we have following definition, and can immediately observe that both $RS$ and $RE$ are label-invariant.


\begin{definition}[Label-invariance]
\label{def:li}
A choice representation $c$ is {\bf label-invariant} if for all $\sigma,\pi \in S_{|U|}$:
\[
c(\sigma\pi) = \{(\pi^{-1}(i),\pi^{-1}(S)) : (i,S) \in c(\sigma) \},
\]
where $\pi^{-1}(S) = \{\pi^{-1}(j)\}_{j \in S}$. 
\end{definition}

\begin{proposition}
\label{prop:RS-label-inv}
The repeated selection ($RS$) and repeated elimination ($RE$) choice representations are label-invariant.
\end{proposition}

\proof
We give the proof for repeated selection; the proof for repeated elimination is equivalent up to the substitution of definitions. Note that for permutations, where composition order matters, $(\sigma \pi)^{-1} = \pi^{-1}\sigma^{-1}$.
\begin{align*}
RS(\sigma\pi) &= \{(\sigma\pi)^{-1}(k), \{(\sigma\pi)^{-1}(k)\}_{j\geq k}\}_{k=1}^{n-1} \\
&= \{(\pi^{-1}(\sigma^{-1}(k)),\{\pi^{-1}(\sigma^{-1}(j))\}_{j\geq k})\}_{k=1}^{n-1}\\
&= \{(\pi^{-1}(x),\pi^{-1}(S)) : (x,S) \in c(\sigma)\}. 
\end{align*}
\endproof

To provide a simple example of a choice representation that is {\it not} label-invariant, suppose we are ranking $n$ items, $U$, and let $c(\sigma) = \{(1,U)\}$ for all $\sigma \in S_n$. This $c$, which is constant for all inputs $\sigma$, is not label-invariant. To see this, let $e$ be the identify permutation and notice that $c(e) = \{(1,U)\}$. We can verify that $c$ is not invariant by considering any $\sigma$ such that $\sigma^{-1}(1) \neq 1$, as $\{(1,U)\} = c(\sigma) = c(e \sigma) \neq \{(\sigma^{-1}(1), U)\}$. This example of a non-label-invariant choice representation is quite pathological, and we are not aware of any plausibly useful choice representations that lack label-invariance. As such, we take label-invariance to be a practically mild assumption.

Label-invariance for choice representations is inspired by the label-invariance property of distances $d$ on $S_n$, sometimes called ``right invariance'' or ``shift invariance,'' whereby $d(\sigma,\tau) = d(\sigma\pi,\tau\pi)$ for all $\sigma,\pi,\tau \in S_n$ \citep{diaconis1988group}. The property has a succinct justification that aligns with our justification for label-invariance: a distance between two rankings should be the same if we relabel the items with a permutation $\pi$.  
The notion of label-invariance is also closely related to other concepts for distance functions on $S_n$ such as $L$-decomposability, $R$-decomposability, and reversibility. We translate several of these related concepts (e.g.,~defining reversibility as a property of a choice model) in Appendix~\ref{app:reverse}.

While label-invariance is a seemingly obvious property of any non-pathological choice representation, it is a powerful enough property to provide the following theorem.
 
\begin{theorem}
\label{thm:Z=1}
For label-invariant choice representation $c$, $Z(p,c) = 1$ for all choice models $p$ if 
there exists some permutation $\pi$ such that $c(\sigma) = RS(\sigma \pi)$. In particular, $Z(p,RS)=Z(p,RE)=1$ for all choice models $p$. 
\end{theorem} 

\proof
Suppose WLOG that we are ranking $U=[n]$. We begin by considering the $RS$ choice representation and showing that for any choice model $p$, 
\[
\sum_{\sigma  \in S_n} \prod_{(i,S) \in RS(\sigma)} p(i,S) =1.
\]

We proceed by induction on $n$, where $n=2$ is trivial:
\begin{eqnarray*}
\sum_{\sigma  \in S_2} \prod_{(i,S) \in RS(\sigma)} p(i,S)
= p(1,\{1,2\}) + p(2,\{1,2\}) = 1.
\end{eqnarray*}

 Assuming that the statement holds for rankings of $n-1$ items, we partition $S_n$ into $A_j = \{\sigma : \sigma(j) = 1\}$ for $j \in [n]$. Then
\begin{align*}
\sum_{\sigma \in S_n} \sum_{(i,S) \in RS(\sigma)} p(i,S)  
&= \sum_{j \in U} \sum_{\sigma \in A_j} \prod_{(i,S) \in RS(\sigma)} p(i,S) \\
&= \sum_{j \in U} \sum_{\sigma \in A_j}  p(j,U) \cdot \left [ \prod_{(i,S) \in RS(\sigma)-(j,U)} p(i,S) \right ]\\
&= \sum_{j \in U} p(j,U) \sum_{\sigma \in S_{|U- j|}} \prod_{(i,S) \in RS( \sigma)}p(i,S)\\
&= \sum_{j \in U} p(j,U) \\
 &=1.
\end{align*}
In the above derivation, the second equality follows from the expansion of the $RS$ representation, and the second to last equality follows from the inductive hypothesis, noting that $|U- j| = n-1$. 

Now consider any permutation $\pi \in S_n$ and the choice representation $c^\pi$ corresponding to selecting the entries of $\sigma$ sequentially in the order given by $\pi$,
$$c^\pi(\sigma) = \{(\sigma^{-1}(\pi^{-1}(i)),\{\sigma^{-1}(\pi^{-1}(j))\}_{j\geq i})\}_{i=1}^{n-1}.$$ 
By label-invariance we have that $c^\pi(\sigma\pi) = RS(\sigma)$ and $c^\pi(\sigma)=RS(\sigma\pi^{-1})$. 
Because composition with $\pi^{-1}$ is an automorphism of $S_n$, we have for any function $f:S_n \ra \R$ that $\sum_\sigma f(\sigma) = \sum_\sigma f(\sigma \pi^{-1})$. 
Taking $f(\sigma) := \prod_{(i,S) \in RS(\sigma)}p(i,S),$
we thus have
\[
\sum_{\sigma} \prod_{(i,S) \in c^\pi(\sigma)} p(i,S) = \sum_{\sigma} f(\sigma \pi^{-1}) = \sum_{\sigma} f(\sigma) =1,
\]
as desired for repeated selection ($RS$).

Letting $\pi$ be the permutation that reverses $\sigma$ gives the analogous desired result for repeated elimination ($RE$). 
\endproof

When $\pi$ is the identity permutation, we have that the normalization constant for $RS$ is always 1, and when $\pi$ reverses its inputs, we have that the normalization constant for $RE$ is always 1. The remaining $\pi$ describe choice models where we choose the $\pi^{-1}(1)$-th position from $U$ first, then the $\pi^{-1}(2)$-th position from what remains, and so forth. As a result, we can conclude that $RS$ and $RE$ are the two ``natural'' choice representations with this property. 

When ranking a prime number of items, the converse statement of Theorem~\ref{thm:Z=1} also holds.

\begin{theorem}
Suppose $n$ is prime and choice representation $c$ is label-invariant and has $Z(p,c)=1$ for any choice models $p$. Then $c(\sigma) = RS(\sigma \tau)$ for some $\tau \in S_n$. 
\label{thm:Z=1revprime}
\end{theorem}

The proof of Theorem~\ref{thm:Z=1revprime} and supporting lemmas appears in Appendix~\ref{app:proofs}. We conjecture that it holds for all $n \geq 2$. In combination with Theorem \ref{thm:Z=1}, these theorems give a powerful argument for using relabelings of repeated selection as choice representations: they are the only choice representations (for prime $n$) guaranteed to be unit normalized regardless of what choice model or family of choice models is considered, greatly simplifying standard methods for inference such as maximum likelihood estimation. 

\subsection{Pairwise choice representation}
\label{sec:pairltr}

While Theorem~\ref{thm:Z=1} and Theorem~\ref{thm:Z=1revprime} guarantee us that choice representations of the form $RS(\sigma \tau)$ for some $\tau \in S_n$ are the only label-invariant choice representations that are unit normalized for every choice model (for prime $n$), other choice representations $c$ may still have $Z(c,p)$ constant, especially when $p$ is restricted to a narrow model class of interest.

One important choice representation in the literature~\citep{rajkumar2014statistical,agarwal2016ranking} is the pairwise choice representation $PW$, as discussed earlier, which outputs all of the pairwise preferences implied by a ranking $\sigma$:
\begin{equation}
PW(\sigma) = \cup_{i,j \in [n] : \sigma(i) < \sigma(j)} \{(i,\{i,j\})\}. 
\end{equation}

It is straightforward to verify that $PW$ is label-invariant, so we know for prime $n$ from Theorem~\ref{thm:Z=1revprime} that $Z(PW,p) \neq 1$ for some $p$. Here we demonstrate that for $n \geq 3$ alternatives, $Z(PW,p)$ is not constant in $p$. Our proof comes from simply computing $Z(PW,p)$ for two simple choice models. 

\begin{proposition}
\label{prop:Zcpair}
For $\sigma \in S_n$, let $p_{\sigma}$ be the ``deterministic" choice model $p_\sigma(i,S) = \1[\sigma(i) \leq \sigma(j), \forall j \in S]$ and let $p_{unif}$ be the choice model $p_{unif}(i,S) = \frac{1}{|S|}$. Then for $n \geq 3$, 
\begin{equation}
Z(PW,p_{unif}) \neq Z(PW,p_\sigma).
\end{equation}
\end{proposition}
\proof

To prove the proposition, we'll simply compute both normalizing constants for $PW$. For $p_{unif}$, all ${n \choose 2}$ choice sets have equal probabilities $1/2$:
\begin{align*}
Z(PW,p_{unif}) &= \sum_\tau \prod_{(i,\{i,j\}) \in PW(\sigma)} 1/2 = \sum_\tau 2^{-{n \choose 2}}= \frac{n!}{2^{n \choose 2}}.
\end{align*}

Because $p_\sigma$ always choose the item ranked highest according to $\sigma$, $\prod_{(i,S) \in PW(\tau) }p_\sigma(i,S)$ is one if $\tau = \sigma$ and zero otherwise. It follows that $Z(PW,p_{\sigma}) = 1$ for any $\sigma$. Now $n!/2^{n \choose 2}<1$ for any $n \ge 3$, so we conclude that $Z(PW,p_{unif}) \neq Z(PW,p_\sigma)$.
\endproof

When $Z(c,p_\theta)$ is not constant across a family of choice models $\{p_\theta\}_{\theta \in \Theta}$ of interest, using maximum likelihood to select a choice model can be complicated in that it requires careful consideration of the normalization constant, as we discuss in more detail in section~\ref{sec:learning}. A consequence of Proposition~\ref{prop:Zcpair} above is then that if our family of choice models includes both $p_{unif}$ and some $p_\sigma$, under the pairwise choice representation $PW$ we will not be able to ignore the normalization constant $Z(PW,p)$ in the likelihood, which can make inference much more difficult. 

\subsection{Partial (top-$k$) rankings}
\label{sec:partialltr}
Throughout this work we generally focus on complete rankings, unless otherwise noted. Because data consisting of incomplete ``top-$k$" ratings are commonplace, we provide a natural extension of repeated selection to such partial rankings. We show in a theorem analogous to Theorem~\ref{thm:Z=1} that this extension has a normalization constant of one for top-$k$ rankings for any $k \in [n-1]$ under repeated selection with any choice model $p$. The normalization constant for top-$k$ rankings under other permuted selection processes is not necessarily one, but may still be constant. In the case of repeated elimination for top-$k$ rankings, we show that the normalization constant is ${n \choose k}$ regardless of choice model $p$.

\subsubsection{Repeated Selection for top-$k$ rankings}
\begin{definition}
A {\bf top-}$k$ {\bf ranking} $\sigma$ of universe $U$ with $|U|=n$ is a one-to-one mapping from some subset $S$ of $U$ with size $k$ to $[k]$. We refer to the set of top-$k$ rankings of $n$ items as $S_{k,n}$. 
\end{definition}

Top-$k$ rankings are common in many empirical datasets we consider. Assuming these rankings came from some repeated selection ranking distribution, we can still obtain the first $k$ choices in the choice representation $RS(\sigma)$ given $\sigma(1),\dots,\sigma(k)$ and $U$.

\begin{definition}
{\bf Repeated selection for a top-}$k$ {\bf ranking} $\sigma$ on universe $U$ with $|U|=n$ and $k \in [n-1]$ is the choice representation
\begin{equation}
RS(\sigma) = \{(\sigma^{-1}(i), U-\{\sigma^{-1}(j)\}_{j\geq i})\}_{i=1}^k.
\label{eq:partial-RS}
\end{equation}
\end{definition}

A top-$(n-1)$ ranking has the same choices as the unique full ranking that completes it, and the extension then agrees with our previous definition of repeated selection. More abstractly, we could equivalently characterize this extension as the intersection of $RS(\tilde \sigma)$ for complete lists $\tilde \sigma$ which match $\sigma$'s first $k$ entries. Given that we've extended $RS$ to top-$k$ rankings, we have a natural extension of $RS_p$ from Definition~\ref{def:c-p} for top-$k$ rankings. We call this choice representation $RS_p$ for top-$k$ lists as well, overloading the notation of $RS_p$ for complete lists but noting that the two choice representations take different arguments (rankings from $S_n$ vs.~top-$k$ rankings from $S_{k,n}$).

\begin{definition}
For discrete choice model $p$ over $n$ alternatives $U$, the $RS_p$ distribution over top-$k$ rankings $\sigma$ is given by 
\begin{equation}
P_{RS,p,k}(\sigma)  = \frac{1}{Z_k(p,RS)} \prod_{(i,S) \in RS(\sigma)} p(i,S),
\end{equation}
which defines a top-$k$ ranking distribution up to a {\it normalization constant} $Z_k(p,RS)$: 
\begin{equation}
Z_k(p,RS) = \sum_{\sigma \in S_{k,n}} \prod_{(i,S) \in RS(\sigma)} p(i,S).
\end{equation}
\end{definition}

We also have a natural analog of Theorem~\ref{thm:Z=1} in the specific case of the choice representation $RS_p$. The proof of this theorem, which handles only $RS_p$ and not $RE_p$ or other permuted choice representations, is simpler than for Theorem~\ref{thm:Z=1} and so we include it here.

\begin{theorem}
\label{thm:partial-RS}
For any choice model $p$ on $n$ alternatives, $Z_k(p,RS) = 1$ for every $k \in [n-1]$. 
\end{theorem}

\proof
We induct on $k$. For $k=1$, we note that a top-$1$ ranking has a single choice $(\sigma^{-1}(1),U)$. Because $p$ is a choice model,  $\sum_{x \in U} p(x,U)=1$ gives $Z_1(p,RS) = 1$.

Assume the proposition holds for $j\leq k$. Recall that $S_{j,n}$ is the set of top-$j$ rankings of $[n]$. For every top-$(k+1)$ ranking $\sigma$ in $S_{k+1,n}$ there exists some unique top-$k$ ranking $\sigma'$ in $S_{k,n}$ such that $\sigma'$ is the prefix of $\sigma$. Furthermore, $RS(\sigma) = RS(\sigma') \cup \{(\sigma^{-1}(k), U - \{\sigma^{-1}(j)\}_{j <k})\}$, and each of the $n-k+1$ rankings $\sigma$ which share $\sigma'$ as a prefix is represented uniquely by $\sigma^{-1}(k)$ being chosen from $U- \{\sigma^{-1}(j)\}_{j <k})\},$ those items unranked by $\sigma'$ . Thus we have
\[
\sum_{\sigma \in S_{k+1,n}} \prod_{(i,S) \in RS(\sigma)}  p(i,S) = \sum_{\sigma' \in S_{k,n}} \sum_{x \in U  - \{\sigma'^{-1}(j)\}_{j<k}} \left ( p(x, U - \{\sigma'^{-1}(j)\}_{j <k}\}) \prod_{(i,S) \in RS(\sigma')} p(i,S) \right ).
\]
Noting that we can pull the product out of the inner sum and that 
$$ 
\sum_{x \in U  - \{\sigma'^{-1}(j)\}_{j<k}}p(x, U - \{\sigma^{-1}(j)\}_{j <k}\}) =1,
$$
because $p$ is a choice model, we obtain
\[
\sum_{\sigma \in S_{k+1,n}} \prod_{(i,S) \in RS(\sigma)}  p(i,S) = \sum_{\sigma' \in S_{k,n}} \prod_{(i,S) \in RS(\sigma')} p(i,S),
\]
and conclude that the sum is 1 for top-$k$ rankings by the inductive hypothesis.  
\endproof

\subsubsection{Repeated Elimination for top-$k$ rankings}
A natural analogous extension of $RE$ to top-$k$ rankings $\sigma$ is to again simply include all of the $RE$ choices that would be in $RE(\sigma')$ where $\sigma'$ is a complete ranking that has $\sigma$ as its prefix. Unlike the $RS$ case, this extension suffers from issues with normalization, as the normalization constant is not 1 for $k<n$. Interestingly enough, however, we can still show that the normalization constant is fixed for all choice models $p$, allowing us to write down the probabilities in a simple closed form.

\begin{definition}
For top-$k$ ranking $\sigma \in S_{k,n}$ where $k \in \{2,\dots,n\}$, let 
\[
RE(\sigma) := \{(\sigma^{-1}(i), \{\sigma^{-1}(j)\}_{j\leq i}\}_{i=2}^k.
\]
\end{definition}

Recall that we use $Z_k$ to denote the normalization constant for choice representations of $S_{k,n}$, i.e.~$Z_k(RE,p) = \sum_{\sigma \in S_{k,n}} \prod_{(i,S) \in RE(\sigma)} p(i,S).$ We will show that $Z_k(RE,p) = {n \choose k}$ for all $p$. We being by stating a simple observation that choices in $RE(\sigma)$  for top-$k$ rankings have choice sets of size at most $k$.

\begin{observation}
$Z_k(RE,p)$ only depends on choice probabilities $p(\cdot,S)$ where $|S| \leq k$.
\label{obs:retopksetsize} 
\end{observation}
The observation follows immediately from noting that for $\sigma \in S_{k,n}$ and $(i,S) \in RE(\sigma)$, $|S| \leq k$. We use this observation to prove our main result for top-$k$ repeated elimination. 

\begin{theorem}
For $k \in \{2,\dots,n\}$, $Z_k(RE,p) = {n \choose k}$ for any choice model $p$. 
\label{thm:ZkRE=nchoosek}
\end{theorem}

The proof of Theorem~\ref{thm:ZkRE=nchoosek} appears in the appendix.

Repeated elimination and repeated selection for top-$k$ rankings lead to very different families of distributions on $S_{k,n}$. One consequence of the normalization constant $Z_k(RE,p) = {n \choose k}$ is that the maximum possible probability of a top-$k$ ranking under $RE$ is $1/{n \choose k}$, whereas repeated selection may assign probability 1 to some top-$k$ ranking. A further consequence is that the support of any distribution using RE for top-$k$ rankings must have size at least ${n \choose k}$, whereas it can be as small as one top-$k$ ranking with repeated selection.

We suspect that some version of the reverse directions of Theorem~\ref{thm:partial-RS} and Theorem~\ref{thm:ZkRE=nchoosek} also hold, akin to how Theorem~\ref{thm:Z=1revprime} gives the reverse direction of Theorem~\ref{thm:Z=1} for $n$ prime. A simple translation of the proof of Theorem~\ref{thm:Z=1revprime} to try and reverse Theorem~\ref{thm:partial-RS} ($RS$ for top-$k$ rankings) falls apart at Lemma~\ref{lemma:bijection}, as there are $(n-k)!$ rankings consistent with the repeated selection choices in a top-$k$ ranking. 
\subsection{Transforming ranking likelihoods to choice likelihoods}
\label{sec:learning}

 Consider a set $T = \{\sigma_j\}_{j=0}^m$ of rankings of items in a universe $U$. Then the likelihood of the choice model parameters $\theta$ for the ranking data $T$ under choice representation $c$ is:
\begin{align}
\label{eq:CTR-lik}
\Li_{c}(\theta; T) = \prod_{\sigma \in T} \frac{1}{Z(p_\theta,c)} \prod_{(i,S) \in c(\sigma)} p_\theta(i,S),
\end{align}
where $Z(p_\theta,c)$ is the normalization constant from Equation~\eqref{eq:z}. We place the normalization constant inside the outer product, highlighting the fact that when $T$ is a set of mixed-length top-$k$ lists (for variable $k$), the normalization constant can depend on $k=len(\sigma)$:

\begin{align}
\label{eq:CTR-lik2}
\Li_{c}(\theta; T) = \prod_{\sigma \in T} \frac{1}{Z_{len(\sigma)}(p_\theta,c)} \prod_{(i,S) \in c(\sigma)} p_\theta(i,S).
\end{align}

The core of our choice-based ranking framework is that, for choice representations that satisfy Theorem~\ref{thm:Z=1} (e.g., repeated selection and repeated elimination), $Z(p_\theta,c)=1$ and we can fit ranking models simply by fitting choice models. The likelihood of the ranking data (above) equals the likelihood of the corresponding choice data $\cup_{\sigma \in T} c(\sigma) $ for the underlying choice model $p_\theta$. 

To see the above claim clearly, consider a set of choices $A=\{(i_k,S_k)\}_{k=0}^m$ (e.g.,~from a choice representation of ranking data, but not necessarily) and a parametric family of choice models $\{p_\theta\}_{\theta \in \Theta}$. We use the notation 
\[
\Li(\theta;A) = \prod_{(i,S) \in A} p_{\theta}(i,S),
\]
to denote the likelihood of $\theta$ for the choice data $A$ under that choice model. We then have the following proposition. 

\begin{proposition}
\label{prop:lik}
For a parametric family of choice models $\{p_{\theta}\}_{\theta \in \Theta}$ and a choice representation $c$ that is repeated selection or repeated elimination, let $\Li_c(\theta;T)$ be the likelihood of a set of rankings $T$ under the choice representation $c$ and choice model $p_\theta$,and let $\Li(\theta; A)$ be the likelihood for a set of choices $A$. Then  
\[
\Li_c(\theta; T) = \Li(\theta; \cup_{\sigma \in T} c(\sigma)).
\]
\end{proposition}

\proof
We have from Theorem~\ref{thm:Z=1} that when $c$ is repeated selection or repeated selection, $Z(p_\theta,c)=\sum_\sigma \prod_{(i,S) \in c(\sigma)} p_\theta(i,S)=1$ for all $\theta \in \Theta$. The rest of the proof is simply reindexing with definitions: the definitions of $\Li$ for sets of rankings and choices and from the definition $c(T):= \cup_{\sigma \in T} \cup_{(i,S) \in c(\sigma)} (i,S)$
\[
\Li_c(\theta;T) = \prod_{\sigma \in T} \prod_{(i,S) \in c(\sigma)} p_\theta(i,S) = \prod_{(i,S) \in \cup_{\sigma \in T} c(\sigma)} p_\theta(i,S) 
= \Li(\theta; \cup_{\sigma \in T} c(\sigma)). 
\]
\endproof

Notice that for choice representations not covered by Theorem~\ref{thm:Z=1}, Equation~\eqref{eq:CTR-lik} makes clear that finding a maximum likelihood estimate of $\theta$ requires engaging with the normalization constant that can vary with $\theta$. Thus, Theorem~\ref{thm:Z=1} and Theorem~\ref{thm:Z=1revprime} are foundational to our choice-based approach to modeling ranking. Because the likelihood of a repeated selection-based distribution is equivalent to the likelihood of the underlying choice model on a transformation of the rankings into choices, we can expect maximum likelihood estimation for a repeated selection {\it ranking model} to be tractable whenever maximum likelihood estimation the underlying {\it choice model} is tractable.

When $c$ is repeated selection, this proposition can be extended to sets $T$ of top-$k$ rankings, appealing to Theorem~\ref{thm:partial-RS} where $Z_k(p,RS)=1$ for any $k \in [n-1]$ as well. 

When $c$ is repeated elimination and $T$ again contains top-$k$ rankings, by Theorem~\ref{thm:ZkRE=nchoosek} we can decompose the log likelihood $\ell_{RE}(\theta;T)$ as 
\begin{eqnarray*}
\ell_{RE}(\theta,T) 
&=& \ell(\theta; \cup_{\sigma \in T} RE(\sigma)) - \sum_{\sigma \in T} \log (Z_k(RE,p_\theta)) \\
&=& \ell(\theta; \cup_{\sigma \in T} RE(\sigma)) - \sum_{\sigma \in T} \log{n \choose len(\sigma)}.
\end{eqnarray*}
Noting that $\sum_{\sigma \in T} \log{n \choose len(\sigma)}$ does not depend on $\theta$, we maximize the ranking log likelihood by simply maximizing the choice log likelihood. 

\section{Ranking with repeated selection}
\label{sec:rs}

We now take a detailed look into ranking distributions arising from repeated selection ($RS$) with choice models. We begin by examining the popular Plackett-Luce and Mallows ranking models and the choice models from which they arise with repeated selection. We then apply repeated selection to the PCMC and CDM choice models, yielding flexible models for rankings from the same conceptual vein as the Plackett-Luce model. 

\subsection{MNL and Plackett-Luce}
The MNL model states that the probability of choosing alternative $i$ from choice set $S$ is proportional to some non-negative ``quality scores'' $\gamma_i$ for all $S$. More precisely, 
\begin{equation}
p_\gamma(i,S) = \frac{\gamma_i}{\sum_{j \in S} \gamma_j}.
\label{eq:MNL}
\end{equation}

Luce noted that Plackett's ranking model \citep{plackett1968random}, since called the Plackett-Luce model, was a direct consequence of his choice axiom and his ranking postulate, which defined repeated selection for the MNL model. For the Plackett-Luce ranking model with its underlying MNL choice model parameterized by quality scores $\gamma$, we have:
\begin{equation}
{P_{RS,\gamma}(\sigma)}  = \prod_{i=1}^{n-1} \frac{\gamma_{\sigma^{-1}(i)}}{\sum_{j \geq i} \gamma_{\sigma^{-1}(j)}}.
\label{eq:PL}
\end{equation}

\subsection{Mallows as repeated selection}

The Mallows distribution assigns probabilities to rankings in a manner that decreases exponentially in the number of pairs of alternatives they ``invert'' relative to some reference permutation $\sigma_0$. More precisely, under a Mallows model with concentration parameter $\theta$ and reference permutation $\sigma_0$,
\begin{equation}
p(\sigma;\sigma_0,\theta) = \frac{
e^{-\theta \tau(\sigma,\sigma_0)}
}{
\sum_{\sigma' \in S_n} e^{-\theta \tau(\sigma',\sigma_0)}
},
\label{eq:mallows}
\end{equation}
where 
\begin{equation}
\tau(\sigma,\sigma_0) = \sum_{i,j \in [n] : \sigma_0(i)<\sigma_0(j)} 1(\sigma(i)>\sigma(j))
\label{eq:tau}.
\end{equation}

There is a known choice model for which repeated selection yields the Mallows ranking model that has been used in work on, e.g.,~assortment optimization \citep{desir2016assortment}. Let $\sigma_0 \in S_n$ be a reference permutation and let $\theta \in \mathbb{R}^+$ be a scalar concentration parameter. The probability that $i$ is ranked first among a subset $S$ for this choice model is then exponential in the number of elements in $S$ ranked above $i$ by the reference permutation $\sigma_0$. 

\citet{fligner1986distance} were among the first to note in that the Mallows model can be composed into what we can interpret as a product of choice probabilities: for choice model 
\begin{equation}
p_{\theta,\sigma_0}(i,S) \propto \exp(-\theta \cdot |\{j \in S: \sigma_0(j)<\sigma_0(i)\}|),
\end{equation}
then $\prod_{(i,S) \in RS(\sigma)} p_{\theta,\sigma_0}(i,S)$ is equivalent to the Mallows density in Equation~\eqref{eq:mallows}.

Although the Mallows model has a convenient form for repeated selection, finding the maximum likelihood reference permutation is NP-hard for as few as four input lists, as previously discussed. In our empirical comparisons between $RS$ mappings of choice models later in this work, we approximate the Mallows ranking distribution ``greedily,'' following \citep{qin2010new}, by building the ranking from front to back by choices that minimize inversions.

The Mallows model is part of a larger class of ``distance-based" models that have been studied in-depth in seminal work on rankings~\citep{diaconis1988group, marden1996analyzing}, with Diaconis singling out the use of Kendall's $\tau$ (which yields the Mallows model) as the most practical distance.  All distance-based models face computational challenges akin to that faced by the Mallows model: they all have a discrete parameter space (all use a ``reference ranking" $\sigma_0$) and as such their MLE cannot be found with traditional gradient-based optimization methods. 

\subsection{Repeated selection with the PCMC choice model}

The pairwise choice Markov chain (PCMC) model is a recent discrete choice model where the selection probability of an item $i$ from a set $S$ is the probability mass on $i$ in the stationary distribution of a continuous time Markov Chain whose transition rates $q_{ij}$ parameterize the model \citep{ragain2016pairwise}. For all $S \subseteq U$, $i \in S$, the model says that $\sum_{j \in S - i} p(i,S)q_{ij} = \sum_{j \in S - i} p(j,S)q_{ji}$. The likelihood of choice data under this model can be maximized over its parameter space of non-negative rate matrices $Q$. The PCMC model is not a RUM. 

\begin{figure}
\centering
\includegraphics[scale=0.55]{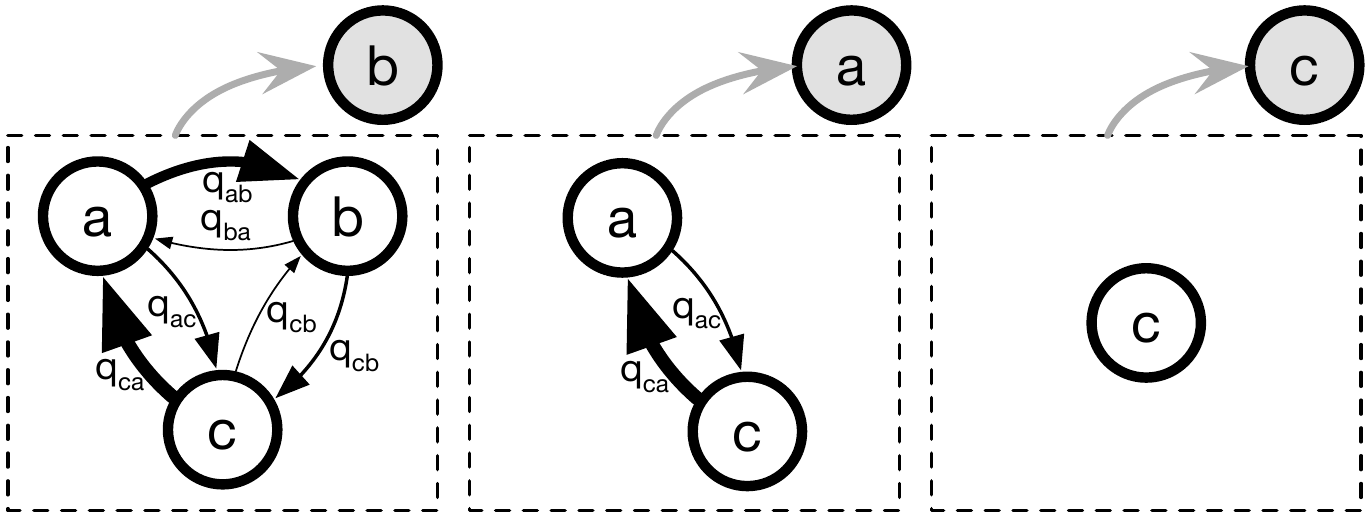}
\caption{Repeated selection from a PCMC model to generate a distribution on rankings. First a choice is made from a CTMC on $\{a,b,c\}$, yielding $\sigma(1)=b$. The state space of the chain is subsetted to the unchosen elements $\{a,c\}$ and a new choice is made, yielding $\sigma(2)=a$. The remaining item is assigned the last position in the ranking, $\sigma(3)=c$.}
\label{fig:diagram}
\end{figure}

Repeated selection from a PCMC model can be conceptualized as a ``darting eye'' process where an individual executes a random walk first on the set of all alternatives and makes a selection according to the stationary distribution of that random walk. The walk then continues over the unused alternatives, mixing fully before making the next selection, and the process continues. For a visualization of this ranking process, see Figure~\ref{fig:diagram}.

The idea of repeatedly applying a Markov chain choice model to a dwindling set of alternatives is importantly distinct from ranking modes based on Markov chains on $S_n$ itself, the space of permutations. Examples of Markov chains on $S_n$ that parameterize ranking distributions include the Move-to-Front \citep{rivest1976self} and Markov Move-to-Front \citep{dobrow1995move} chains. We discuss contrasts between choice-based ranking Markov chain models and $S_n$-based Markov chain models in Appendix~\ref{sec:mc-on-sn}.

We note that the MNL choice model family is a subset of the PCMC choice model family with a specific parameterization of the transition rates: given an MNL model with parameters $\gamma = (\gamma_1,...,\gamma_n)$ and setting $q_{ij} = \gamma_i/(\gamma_i + \gamma_j)$, $\forall i,j$, the stationary distributions of the Markov chains correspond to the choice probabilities of the given MNL model.

\subsection{Repeated selection with the CDM choice model}

The context-dependent model (CDM) is another recently developed model for choice which focuses on contextual utility \citep{cdm}. A $d$-dimensional CDM includes $2dn$ parameters, with each alternative $i \in U$ having a feature vector $a_i \in \R^d$ and a context vector $b_i \in \R^d$. The utility of alternative $i$ in the context of $S$ is simply the sum of the dot product of $i$'s features with each context vector in the choice set. The choice probabilities are proportional to the exponential of these contextual utilities, i.e., 
\begin{equation}
p(i,S) \propto e^{u_{iS}}, \ u_{iS} := \sum_{j \in S}a_i \cdot b_j.
\end{equation}

An equivalent characterization of the CDM model considers pairwise contextual utilities $u_{ij} := a_i\cdot b_j$ as the full-rank parameterization of the model and notes that $u_{iS} = \sum_{j \in S} u_{ij}$. For $d$-dimensional feature and context vectors, it is equivalent to constraining the rank of the matrix of contextual pairwise utilities $\mathcal{U} = \{u_{ij}\}_{i \in U, j \in U}$ (recall that $U$ here is the universe of alternatives). Because the contextual utilities depend on what is in the choice set $S$, the CDM is not a RUM.

As with the PCMC model, the MNL model family is also a subset of the CDM model family. One construction of a CDM model that is equivalent to a given MNL model with parameters 
$\gamma$ is to set $u_{ij}= \log \gamma_j$ for all $i,j \in U$. 

\section{Repeated selection and choice axioms}

One appeal of choice-based ranking is that existing knowledge about a choice model may provide structure to the resulting ranking distribution under that choice model. Here we translate several choice axioms through the repeated selection choice representation. Readers less interested in axiomatic considerations are encouraged to advance to the empirical results presented Section~\ref{sec:empirics}.

\subsection{Luce's choice axiom (IIA)}
Luce's choice axiom \citep{luce1959individual}, also known as the independence of irrelevant alternatives (IIA), states that (i) the probability of choosing an element $i$ from $S$ conditioned on choosing an element from some subset $S'$ of $S$ is equal to the probability of choosing $i$ from $S'$, and (ii) if $i$ is never chosen from $S$, $i$ is never chosen from any set containing $S$. Luce showed that any model satisfying the choice axiom was equivalent to some MNL model, which defines the selection probability of $i$ in $S$ proportional to some latent quality $\gamma_i$ for each choice set $S$ (in Luce's words, every such model admits a {\it ratio scale representation} \citep{luce1977choice}). Thus the repeated selection ranking distribution of any choice model satisfying IIA will be equivalent to some Plackett-Luce distribution with parameters $\gamma$.

\subsection{Local and Nested IIA}

Although celebrated as a seminal choice axiom, IIA often fails to hold for empirical data \citep{benson2016relevance,boyd1980effect}. A popular generalization of MNL that escapes IIA is the Nested Multinomial Logit model (NMNL) that partitions the universe into subsets called nests for which IIA holds \citep{mcfadden1978modeling}.

Recall that we let $p(i,T)$ represent the probability that $i$ is chosen from $T$ and $p(S,T) = \sum_j p(j,T)$ for $S \subseteq T$. If we unpack Luce's axiom with Bayes' rule we see that if $x$ is the chosen element,
\[
Pr(x=i | x \in S) = \frac{Pr(x \in S | x \in T)Pr(x=i | x\in T)}{Pr(x \in S | x \in T)}= \frac{p(i,T)}{p(S,T)},
\]
and from the choice axiom for subset $S$ of $T$ we have $P(x = i | x \in S) = p(i,S)$. We thus have $p(i,T) = p(S,T)p(i,S)$. For the MNL model, this relationship holds for any $S\subseteq T \subseteq U$. For other models it may not hold for all pairs of $S$ and $T$ but may still hold for many pairs in a way that provides instructive structure. The benefit of this formulation is that it allows us to express the axiom entirely with our adopted notation for the choice model, $p$. 

\begin{definition}
For a choice model $p$ on universe $U$ and subsets $S$ and $T$ with $S \subseteq T \subseteq U$, $(p,S,T)$ exhibits {\bf local IIA} if for all $i \in S$:
\begin{equation}
p(i,S) = p(S,T)p(i,T).
\label{eq:local-IIA}
\end{equation}
\end{definition}

While the NMNL model is usually considered for data consisting of a single choice set, this work focuses on repeated selection from a choice model, which will typically use all subsets of $U$ as choice sets. Because of this, we want to focus on choice models where the nesting structure is not destroyed by the removal of some of the alternatives. 
\begin{definition}
For a choice model $p$ on universe $U$ and subsets $S$ and $T$ with $S\subseteq T \subseteq U$, $(p,S,T)$ exhibits {\bf nested IIA} if for all $S' \subseteq S$:
$$
p(i,T-S') = p(S-S',T-S')p(i,S-S'). 
$$
\end{definition}

Both the PCMC model and the CDM model can exhibit nested IIA under reasonable conditions, giving structure to both their choice probabilities and their repeated selection distributions. Proofs of both propositions appear in the appendix. 

\begin{proposition}
\label{prop:PCMC-nest}
For a PCMC choice model $p$ parameterized by rate matrix $Q$ and a partition $S_1,\dots,S_k$ of $U$ where for all $i \in S_I, j \in S_J$ with $I \neq J$, $q_{ij} = \lambda_{IJ}$, then for each $I$, $(p,S_I,U)$ exhibits nested IIA.
\end{proposition}

\begin{proposition}
\label{prop:CDM-nest}
For a CDM choice model $p$ parameterized by pairwise contextual utilities $\mathcal{U}=\mathbf{f}^T\mathbf{c}$ with $\mathcal{U}_{ij}$ fixed for all $i \in S, j \not \in S$, $(p,S,U)$ exhibits nested IIA.
\end{proposition}

Nested IIA identifies sets for which we can easily decompose the repeated selection probabilities into choosing a subset and choosing from within that subset for every choice in the choice representation $RS(\sigma)$. If $(p,S,U)$ exhibits local IIA, then for any $i \in S$ we can decompose the choice probabilities as in Equation~\eqref{eq:local-IIA}, so for the choice probabilities for the first item $\sigma^{-1}(1)$ in a ranking $\sigma$ from repeated selection, the decomposition applies to $p(x,U)$ for $x \in S$. 

Local IIA alone only applies to $p(\cdot,U)$, though, whereas $RS(\sigma)$ contains choices from other choice sets, so we cannot say anything about $P_{RS,p}(\sigma)$ with Local IIA on $U$ alone. Nested IIA gives us that the structure in $p(\cdot,U)$ extends to the subsets of $U$ which appear as choice sets in $RS(\sigma)$, so Nested IIA gives us enough structure to $p$ to make statements about $P_{RS,p}$. 

Consider the second choice in a ranking under in a repeated selection model, choosing $\sigma^{-1}(2)$ from $U-\sigma^{-1}(1)$, where $\sigma^{-1}(1)$ was in some subset $S$ exhibiting local IIA for $U$. If $S$ and $U$ exhibited nested IIA for $p$, then for $i \in S-\sigma^{-1}(1)$, we have 
$$
p(i,U-\sigma^{-1}(1)) = \left(\sum_{j \in S-\sigma^{-1}(i)} p(j,U-\sigma^{-1}(1))\right)p(i,S-\sigma^{-1}(1)),
$$ but if the local IIA was not nested, we cannot say anything about the choice probabilities for $i \in S$ from $U-\sigma^{-1}(1)$ when $\sigma^{-1}(1) \in S$, as the local IIA may have been ``destroyed" when $\sigma^{-1}(1)$ was removed. 

When $(p,S,U)$ further exhibit nested IIA, however, we can apply the decomposition in Equation~\eqref{eq:local-IIA} to $S \cap T$ for every choice set $T$ in $RS(\sigma)$. As a result, for $\sigma$ from $P_{RS,p}$, the relative ordering of the items in $S$ is independent of the set of positions at which they land, $\{\sigma(i)\}_{i \in S}$, and as a result restricting $\sigma$ to alternatives in $S$ yields a repeated selection distribution with $p$ restricted to subsets of $S$. We thus have the follow proposition, a proof of which appears in the appendix. 

\begin{proposition}[$RS$ within $RS$ for nested IIA]
\label{prop:rs-nested-IIA}
For a choice model $p$ on universe $U$, suppose $(p,S,U)$ exhibits nested IIA. For $\sigma$ drawn from $P_{RS,p}$, let $\sigma_S$ be the restriction of $\sigma$ to $S$ and $\tilde p_S$ be the restriction of $p$ to $S$. Then $\sigma_S$ is distributed according to $P_{RS,\tilde p_S}$.
\end{proposition}

\subsection{Regularity}
Regularity for a choice model stipulates that for all $S' \subseteq S$ and all $S \subseteq U$, $p(i,S')\geq p(i,S)$. As a result, $p(i,S) \geq p(i,U)$ for all $i \in S$ for any subset $S$ of $U$, giving us a simple lower bound on the probability of any ranking. This lower bound is most useful when $\min_{i \in U} p(i,U)$ is not too small. 

\begin{proposition}
\label{prop:rs-regularity}
For any $\sigma$ drawn from repeated selection according to a choice model $p$ exhibiting regularity,
\begin{align}
Pr(\sigma) \geq \frac{\prod_{i \in U} p(i,U)}{\max_{i \in U} p(i,U)}.
\end{align}
\end{proposition}
\proof
For every $k \in [n-1]$, 
\[
p(\sigma^{-1}(k), \{\sigma^{-1}(j)\}_{j>k})  \geq \  p(\sigma^{-1}(k),U),
\]
by regularity, as $ \{\sigma^{-1}(j)\}_{j>k} \subseteq U$. Because $\sigma^{-1}$ is a bijection, for each $k \in [n-1]$ we have that $\sigma^{-1}(k)$ is chosen exactly once, and thus
\[
 \prod_{k=1}^{n-1} p(\sigma^{-1}(k),\{\sigma^{-1}(j)\}_{j >k}) \geq \prod_{k=1}^{n-1} p(\sigma^{-1}(k),U) = \prod_{i \in U : \sigma^{-1}(i) < n} p(i,U) \geq \frac{\prod_{i \in U} p(i,U)}{\max_{i \in U} p(i,U)}. 
\]
\endproof

\begin{corollary}
If $\sigma$ is drawn from repeated selection on a choice model $p$ exhibiting regularity and with $p(i,U) \geq \epsilon$ for all $i \in U$ and some $\epsilon > 0$, then $P_{RS,p}(\sigma) \geq \epsilon^{n-1}$. Furthermore, any independent RUM model with random utilities having full support on $\R$ will admit a ranking model through repeated selection with full support on $S_n$. 
\label{cor:rs-regular-epsilon}
\end{corollary}
\proof
This corollary falls out of Proposition~\ref{prop:rs-regularity} simply by applying $p(i,U) \geq \epsilon$, but can also be shown directly. Let $i_{\min}$ be the element of $U$ minimizing $p(i,U)$. Then for any subset $S$ and $i \in S$, by regularity $p(i,S) \geq p(i,U) \geq p(i_{\min},U) \geq \epsilon$. Noting that the last choice in repeated selection is always a probability 1 choice from a singleton, the probability of $\sigma$ is at least the probability of the first $n-1$ choices, all bounded below by $\epsilon$. 
\endproof

Regularity holds for RUMs in particular, giving strong restrictions on the repeated selection ranking distribution based on any RUM. Independent random utilities are typically modeled as continuous distributions with full support, e.g.~Gumbel as in the MNL model or Normal as in the Thurstone model, and thus have strictly positive choice probabilities for each alternative in $U$.

\section{Empirical results}
\label{sec:empirics}

Here we compare the performance of repeated selection and repeated elimination models when training and making predictions on empirical datasets. The datasets span a wide variety of human decision domains including ranked elections and food preferences, while also including (search) rankings made by algorithms. We specifically highlight the performance of the $RS_{PCMC}$ and $RS_{CDM}$ ranking models for their ability to learn complex distributions to represent human ranking data. 

We evaluate predictions based on the negative log-likelihood for out-of-sample ranking data, handled as choice data in its choice representation. 
Choice-based ranking also enables an examination of predictive performance at each position. For repeated selection, we can use our estimated choice models $p_{\hat \theta}(\cdot,S)$ to predict the next entry of each $\sigma \in T$ at each position, given that $S$ is the set of items we haven't yet chosen from $\sigma$.

We can measure the error at the $k$-th position of a ranking $\sigma$ given the set of already ranked items each by adding up some distance between the choice probabilities $p$ for the corresponding choice sets and the empirical distribution of those choices in the data. 
For repeated selection models, we define the {\it position-level log-likelihood} at each position $k$ as
$
\ell(k, \theta; \sigma) := \log p_{\theta} (\sigma^{-1}(k), \{\sigma^{-1}(j)\}_{j\geq k}).
$
When averaging $\ell$ over a test set $T$ we obtain the average position-level log-likelihood: 
\begin{eqnarray}
\ell(k; \theta, T) := \frac{1}{|T|} \sum_{\sigma \in T : len(\sigma)\geq k} \ell (k, \theta ; \sigma),
\label{eq:logcsr2}
\end{eqnarray}
where $len(\sigma)$ is $n$ for a full ranking and $k$ for a top-$k$ ranking. 

For mixed length top-$k$ rankings represented by repeated elimination, in order to evaluate choice models based on the likelihood of out-of-sample test data we must take into consideration the heterogeneous normalization constants $Z_k(RE,p) = {n \choose k}$. Because the product choice probabilities alone do not entail the top-$k$ ranking probabilities, position-level log-likelihoods do not translate to a straightforward way to repeated elimination.

\subsection{Datasets}
We consider a wide variety of application domains to demonstrate that repeated selection and repeated elimination are appropriate and practical for general ranking problems. Many of our datasets can be found in the Preflib library\footnote{Preflib data is available at: \href{http://www.preflib.org/}{http://www.preflib.org/}}.

In our evaluation we place a particular emphasis on four widely studied collections of human choices. First, the {\tt sushi} dataset, consisting of 5,000 complete rankings of 10 types of sushi. 
Next, three election datasets, which consists of ranked choice votes given for three 2002 elections in Irish cities: the {\tt dublin-north} election (abbreviated {\tt dub-n} in tables) had 12 candidates and 43,942 votes for lists of varying length, {\tt meath} had 14 candidates and 64,081 votes, and {\tt dublin-west} (abbreviated {\tt dub-w}) had 9 candidates and 29,988 votes. 
Third, the {\tt nascar} dataset representing competitions, which consists of the partial ordering given by finishing drivers in each race of the 2002 Winston Cup. The data includes 74 drivers (alternatives) and 35 races (rankings). 

The fourth collection we emphasize is the popular {\tt LETOR} collection of datasets, which consists of ranking data arising from search engines. Although the {\tt LETOR} data arises from algorithmic rather than human choices, it demonstrates the efficacy of our algorithms in large sparse data regimes. After removing datasets with fewer than 10 rankings and more than 100 alternatives (arbitrary thresholds that exclude small datasets with huge computational costs), the {\tt LETOR} collection includes 727 datasets with a total of 12,838 rankings of between 3 and 50 alternatives.

Beyond these four emphasized collections, we include analyses of all 51 other Preflib datasets (as of November 2018) that contain partial or complete rankings of up to 10 items and at most 1000 rankings, a total of 11,956 rankings (these thresholds were again decided arbitrarily for computational reasons). 
We call this collection of datasets {\tt PREF-SOI}, adopting the notation of \citet{MaWa13a}. 
We separately study the subset of 10 datasets comprised of complete rankings, referred to here-in as {\tt PREF-SOC}, which contain a total of 5,116 rankings. The complete rankings in the {\tt PREF-SOC} collection are suitable for both repeated selection and repeated elimination. 
While the (complete ranking) sushi and (partial ranking) election datasets are part of Preflib, they are comparatively quite large and are excluded from these two collections ({\tt PREF-SOC} and {\tt PREF-SOI}, respectively) by the above thresholds. 

\subsection{Training}

We use the stochastic gradient-based optimization method Adam \citep{kingma2014adam} implemented in Pytorch to train the MNL, PCMC, and CDM-based models in this paper. We run Adam with the default parameters ($lr=0.001$, $\beta= (0.9, 0.999)$, $\epsilon=1e-8$). We use $10$ epochs of training except for the {\tt election} datasets, where a single epoch converged. 

We cannot use Adam (or any simple gradient-based method), for the Mallows model as the reference permutation parameter $\sigma_0$ lives in a discrete space. Instead we select the reference permutation via the Mallows Greedy Approximation (MGA) as in \citep{qin2010new}. Given a reference permutation, the concentration parameter $\theta$ has a tidy maximum likelihood estimate, see Appendix~\ref{sec:mallows-mle}. Our results broadly show that the Mallows model performs poorly compared to all the other models, including even the uniform distribution (a naive baseline), so we exclude it from some of the more detailed evaluations. 

For all datasets we use 5-fold cross validation for evaluating test metrics. Using the {\tt sushi} dataset as an example, for each choice model we train on $RS$ and $RE$ choice representations for each of 5 folds of the 5,000 rankings in the dataset. Across 5 choice models, this amounts to 10 fits (both $RS$ and $RE$) on 5 folds of 5,000 rankings, represents training on a total of 200,000 rankings that represent 2 million choices. The training process can be easily guided to exploit sparsity, parallelization, and batching. Repeated elimination ($RE$) models are trained analogously to repeated selection ($RS$) models, simply reversing the training rankings. All replication code is made available.\footnote{\href{https://github.com/sragain/CTR}{https://github.com/sragain/CTR}.}

\subsection{Log-likelihood for ranking data}

In Table~\ref{table:rs-all} we report out-of-sample log-likelihood for all the datasets and collections of datasets we study. On the {\tt sushi} dataset we find that repeated selection with the one-dimensional CDM-based and the PCMC-based model offer slight improvements over the MNL-based model, while repeated selection with higher-dimensional CDM-based models offer significant additional improvements. We see a similar pattern for all three election datasets, {\tt dublin-north}, {\tt dublin-west}, and {\tt meath}. For all datasets, the Mallows Greedy Approximation (MGA)-based model is markedly worse than the other models.

For the {\tt nascar} dataset, we find that the CDM-based models and MNL-based models are relatively equal, while repeated selection with the PCMC-based model preforms significantly better. There are more alternatives/racers in this dataset than the others, but few rankings/races, so the PCMC-based model appears to be benefitting from its larger number of parameters to represent this complex data (note that all results are out-of-sample). Another more remote possibility is that the default parameters for Adam somehow suit the $RS_{PCMC}$ models better than the $RS_{CDM}$ models. Brief explorations gave no indication that this is the case, as other Adam parameter values did not meaningfully change the relative performance of these methods. 

Similarly, we find for the {\tt LETOR} collection of datasets that the CDM-based models perform roughly as well as the MNL-based model across all dimensions, but we see that the PCMC-based model performs significantly better. The dip in out-of-sample performance for the eight-dimensional CDM relative to the four-dimensional CDM may suggest that $RS_{CDM}$ begins to overfit the data when given more parameters. $RS_{PCMC}$, which has many still more parameters, does not appear to have this problem.

 \begin{table}
\small
\centering
\begin{tabular}{ c | c c c c c c }
  & $RS_{MNL}$ & $RS_{CDM,d=1}$& $RS_{CDM,d=4}$& $RS_{CDM,d=8}$& $RS_{PCMC}$& $RS_{MGA}$ \\ 
 \hline
{\tt sushi} &14.24 $\pm$ 0.02 &13.94 $\pm$ 0.02 &13.57 $\pm$ 0.02 &{\bf13.47} $\pm$ 0.02 &13.91 $\pm$ 0.02 &34.94  $\pm$ 0.06 \\
{\tt dub-n} &8.36 $\pm$ 0.02 &8.18 $\pm$ 0.02 &7.61 $\pm$ 0.02 &{\bf 7.59} $\pm$ 0.02  &8.15 $\pm$ 0.02  & 28.70  $\pm$ 0.05  \\
{\tt dub-w} &6.36 $\pm$ 0.02&6.27  $\pm$ 0.02 & 5.87 $\pm$ 0.02 &{\bf  5.86}  $\pm$ 0.01&6.12 $\pm$ 0.02&16.88   $\pm$ 0.07\\
{\tt meath}& 8.46  $\pm$ 0.02&8.23  $\pm$ 0.02 & 7.59 $\pm$ 0.02 &{\bf 7.56}  $\pm$ 0.02& 8.05 $\pm$ 0.02 & 34.31   $\pm$ 0.07\\
{\tt nascar} &113.0  $\pm$ 1.4 & 112.1 $\pm$ 1.5 & 103.9 $\pm$ 1.8 & 102.6  $\pm$ 1.8 & {\bf 95.6}  $\pm$ 1.2 & 3485  $\pm$ 7.2\\
{\tt LETOR} &12.2 $\pm$ 1.0 &12.2 $\pm$ 1.0 &10.5 $\pm$ 1.1 & 9.8  $\pm$ 1.1 &{\bf 9.3}  $\pm$ 0.8 &   92.4 $\pm$ 3.1\\
{\tt PREF-SOC} & 5.52  $\pm$ 0.08 & 5.53  $\pm$ 0.07 & 5.55 $\pm$ 0.14 & 5.54 $\pm$ 0.15 & {\bf 5.20} $\pm$ 0.10 &  16.7 $\pm$ 2.3 \\
{\tt PREF-SOI} & 4.1  $\pm$ 0.1 & 4.0  $\pm$ 0.1 & {\bf 3.9}  $\pm$ 0.1 & {\bf 3.9}  $\pm$ 0.1  & {\bf 3.9}  $\pm$ 0.1  & 10.22 $\pm$ 0.4  \\
\end{tabular}
\caption{Mean out-of-sample negative log-likelihood for the MLE of repeated selection ($RS$) representations across different datasets (lowercase names) or collections of datasets (uppercase names), $\pm$ standard errors (of the mean). The best result for each dataset appears in bold. Results are averaged across five folds.}
\label{table:rs-all}
\end{table}

 \begin{table}
\small
\centering
\begin{tabular}{ c | c c c c c c }
  & $RE_{MNL}$ & $RE_{CDM,d=1}$& $RE_{CDM,d=4}$& $RE_{CDM,d=8}$& $RE_{PCMC}$ & $RE_{MGA}$\\ 
 \hline
{\tt sushi} & 14.13  $\pm$ 0.02 & 13.97 $\pm$ 0.03 & 13.53 $\pm$ 0.02 & {\bf13.49} $\pm$ 0.03 & 13.94 $\pm$ 0.03& 35.4 $\pm$ 0.1 \\
{\tt dub-n} & 8.69 $ \pm $ 0.02 & 8.57 $ \pm $ 0.02 & 8.24 $ \pm $ 0.02 & {\bf 8.22} $ \pm $ 0.02 & 14.86 $ \pm $ 0.03 & 9.88 $\pm$ 0.005   \\
{\tt dub-w} &6.60 $\pm$ 0.002 & 6.56 $\pm$ 0.02 & 6.39 $\pm$ 0.02 & {\bf 6.37} $\pm$ 0.02 & 11.43 $\pm$ 0.02& 7.26 $\pm$ 0.008\\
{\tt meath} & 8.71 $ \pm $ 0.02 & 8.49 $ \pm $ 0.02 & 8.09 $ \pm $ 0.02 & {\bf 8.05} $ \pm $ 0.02 & 15.51 $ \pm $ 0.03& 9.98 $\pm$ 0.005 \\
{\tt nascar} &119.0  $\pm$ 2.1 & 112.6  $\pm$ 1.4 & 105.4 $\pm$ 1.7 & {\bf 97.5}  $\pm$ 1.7 & 170.8  $\pm$ 1.4 & 665.6 $\pm$ 4.7\\
{\tt LETOR} & 12.8 $\pm$ 1.0 & 12.4.$\pm$ 1.0 & 10.6 $\pm$ 1.0 & {\bf 9.7} $\pm$ 1.0 & 20.5 $\pm$ 1.5 & 34.4 $\pm$ 1.0  \\
{\tt PREF-SOC} & 5.59  $\pm$ 0.06 & 5.56  $\pm$ 0.08 & 5.29 $\pm$ 0.10 &{\bf 5.17} $\pm$ 0.11 & 6.13 $\pm$ 0.15 & 16.7 $\pm$ 2.3 \\
{\tt PREF-SOI} & 4.2  $\pm$ 0.1 & 4.1 $\pm$ 0.1 & {\bf 4.0} $\pm$ 0.1 & {\bf 4.0}  $\pm$ 0.1  & 7.4  $\pm$ 0.2 & 5.8 $\pm$ 0.2 \\
\end{tabular}
\caption{Mean out-of-sample negative log-likelihood for the MLE of repeated elimination ($RE$) representations across different datasets (lowercase names) or collections of datasets (uppercase names), $\pm$ standard errors (of the mean). The best result for each dataset appears in bold. Results are averaged across five folds.}
\label{table:re-all}
\end{table}


Table~\ref{table:re-all} gives the out-of-sample log-likelihood for repeated elimination models for all of the datasets. Comparing Tables~\ref{table:rs-all} and \ref{table:re-all}, we find that the repeated selection and repeated elimination models generally perform quite comparably for MNL-based and CDM-based models, though $RE$ models tend to perform slightly worse than $RS$ models, except on the {\tt sushi} dataset.  Noting that the {\tt sushi} data has previously been distributed and analyzed in prior work in the ``wrong'' order~\citep{sushiwarning}, it is interesting that the data is in fact slightly more predictable (under $RS_{MNL}$, which is Plackett-Luce) in the wrong/reversed order than in the correct order.  The performance difference between $RS_{PCMC}$ and $RE_{PCMC}$ is much more noticeable, with $RE_{PCMC}$ performing worse than all other $RE$ models. While $RE_{MGA}$ results are much worse than other $RE$-based results, they are not as bad as $RS_{MGA}$. We think that the $RS_{MGA}$ distributions perform poorly in part due to the thin ``tails" of the distribution, which do not fit messy human data well. This intuition further supported by the better performance of $RE_{MGA}$ for top-$k$ lists (with $k < n$), where the modeling is confined to smaller sets. 

\subsection{Position-level evaluation results}

\begin{figure}[t]
\centering
\includegraphics[width=0.49\columnwidth]{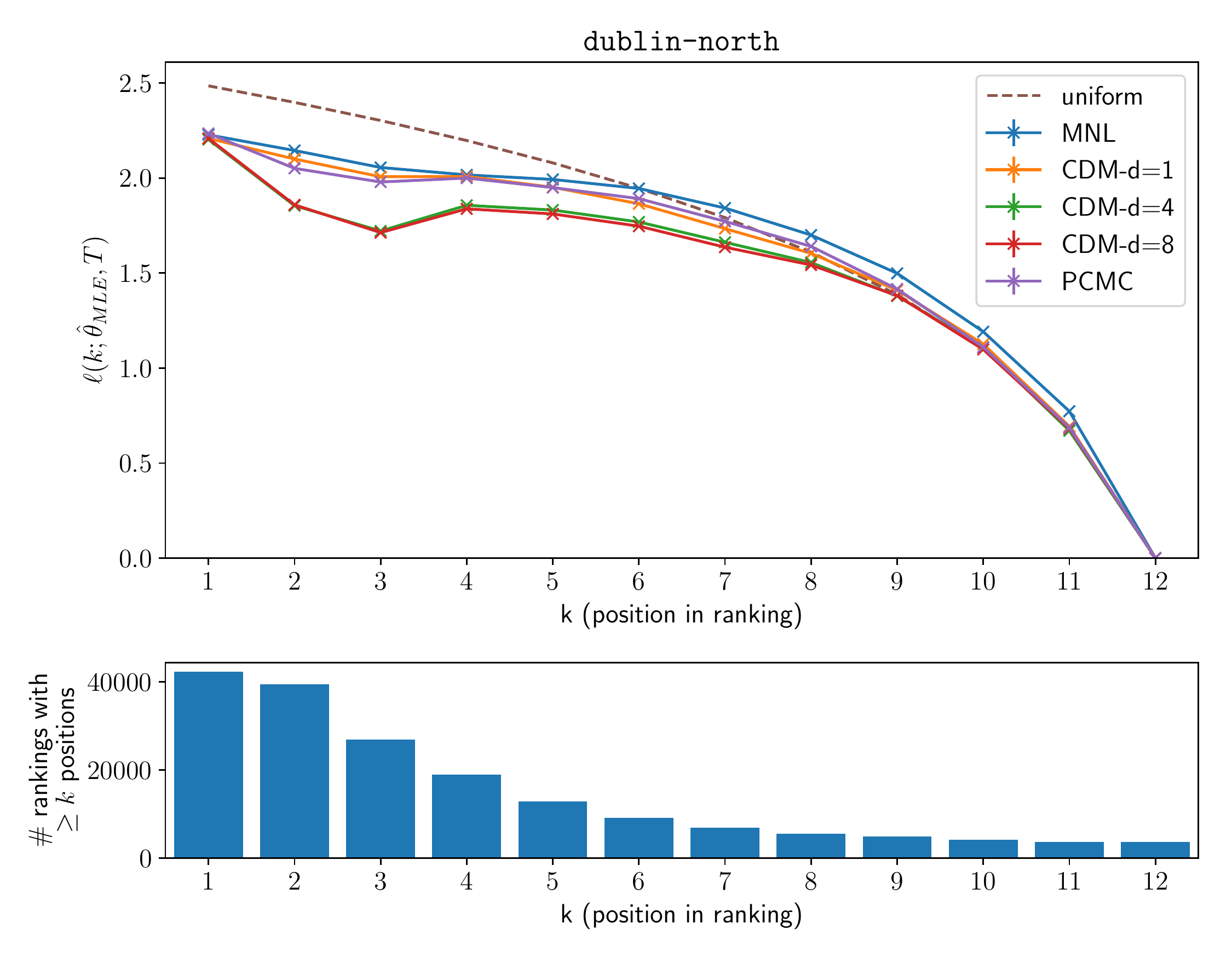}
\includegraphics[width=0.49\columnwidth]{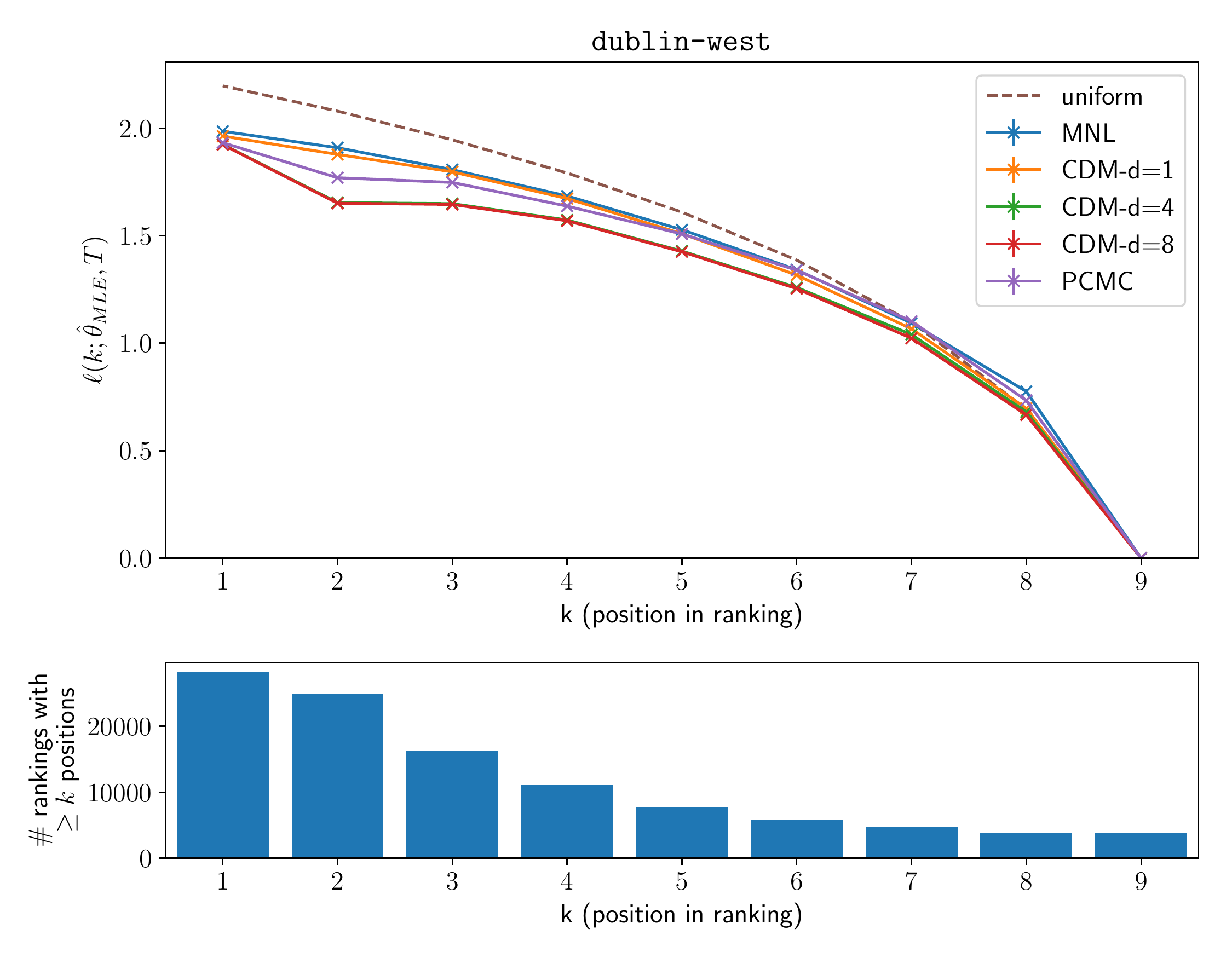}
\includegraphics[width=0.49\columnwidth]{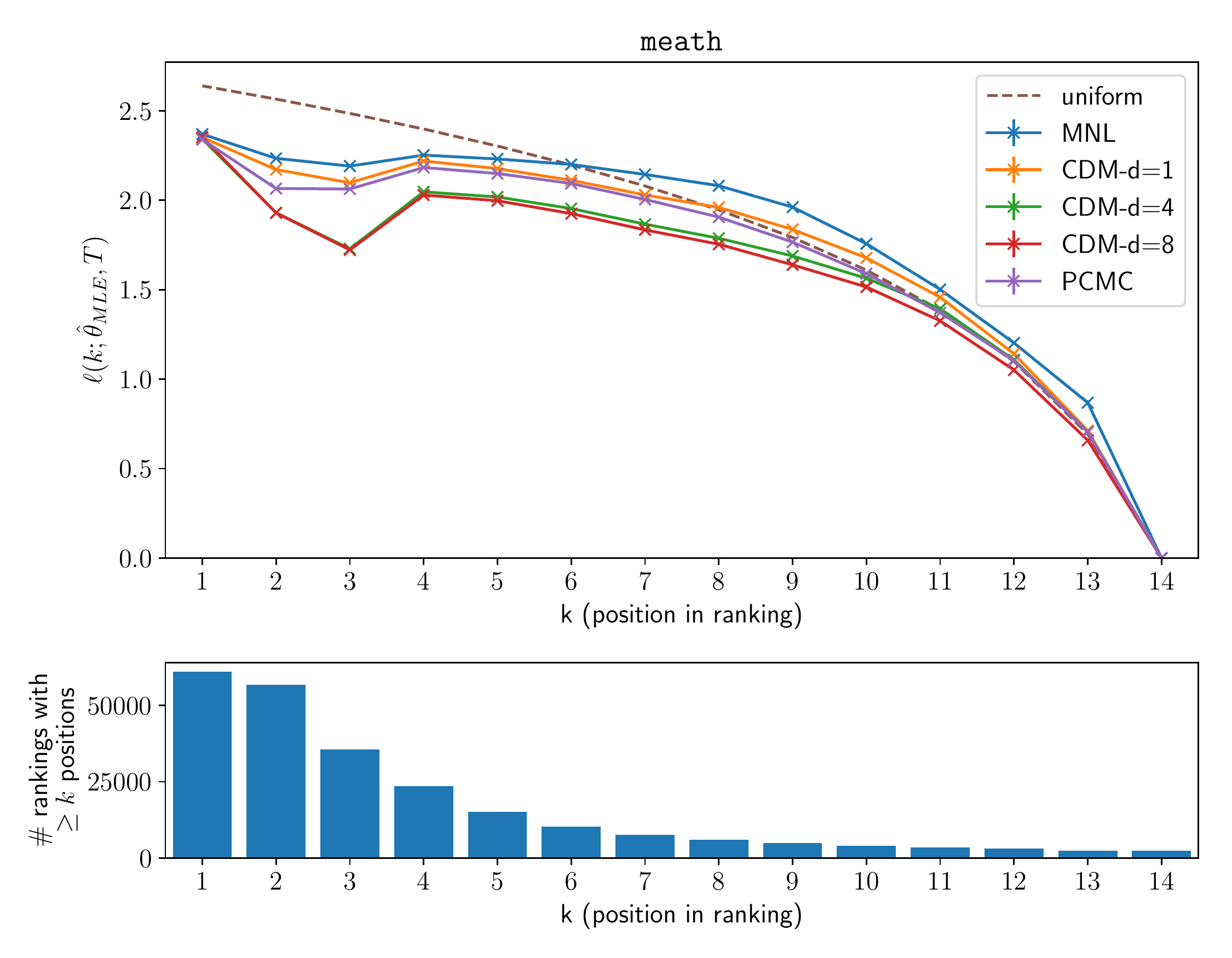}
\includegraphics[width=0.49\columnwidth]{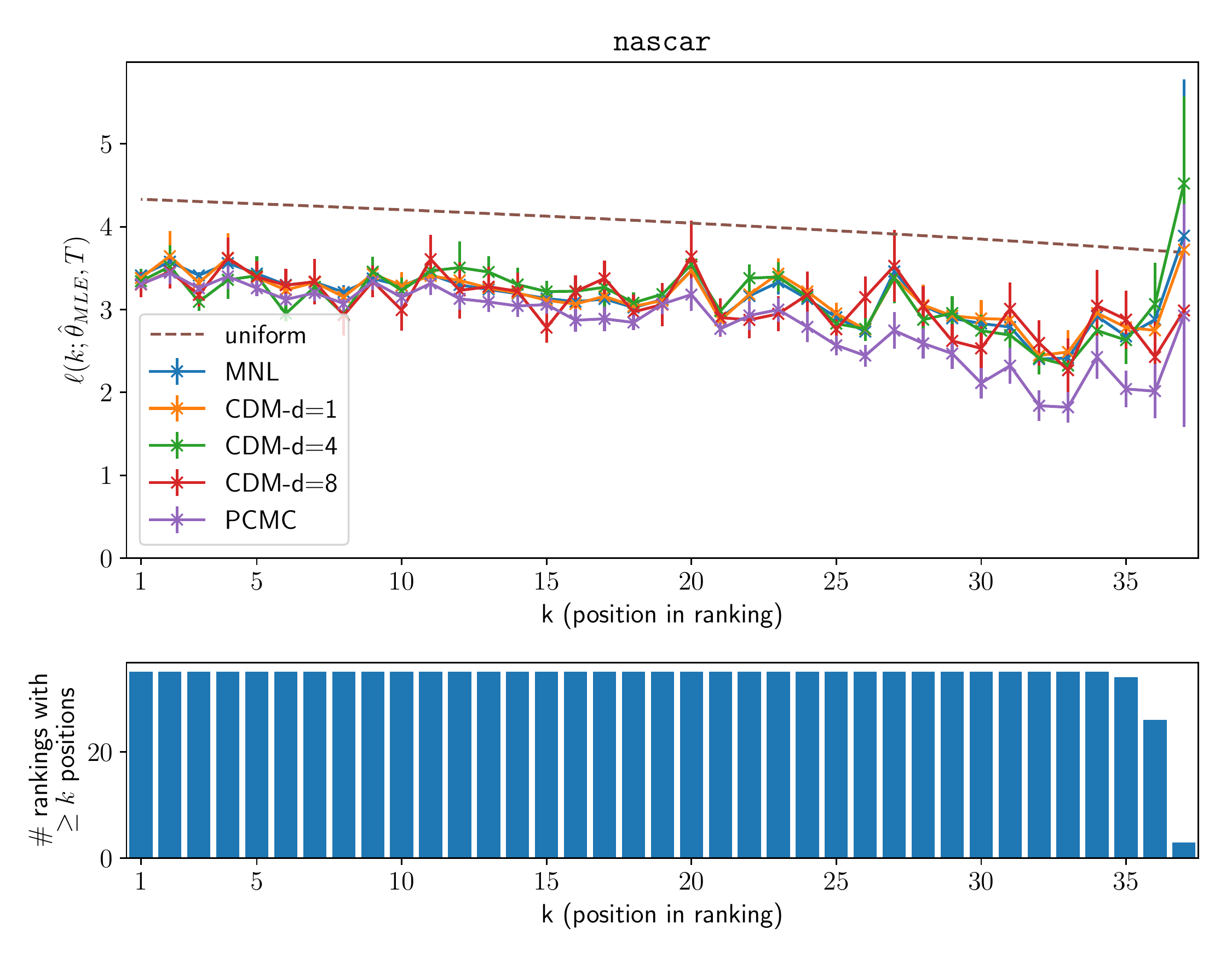}
\caption{The average position-level log likelihood of choice probabilities and accompanying histograms of ranking lengths for the {\tt dublin-north} (top left), {\tt dublin-west} (top right), {\tt meath} (bottom right), and {\tt nascar} (bottom right) datasets. }
\label{fig:logscr-plots}
\end{figure}

In order to analyze the differences between choice-based ranking models, we examine the log-likelihood of the choices at individual positions, as defined in Equation~\eqref{eq:logcsr2}, for the election, {\tt nascar}, and {\tt sushi} datasets. For repeated selection we look at the log probability of the choices at each position, reading each ranking from top to bottom as we reveal more of the prefix. For repeated elimination (of complete rankings) we look at the log probability of the entries reading from bottom to top as we reveal more of the suffix. Our plots include error bars denoting standard errors, though the error bars are typically very small because there are many test lists when averaging over the data.

In Figure~\ref{fig:logscr-plots} we analyze the election datasets, where we find that more nuanced choice models make significant gains relative to simpler models when predicting candidates near---but not at---the top of the list. We further notice that for both the CDM and PCMC-based models  the performance is not monotonically decreasing in the number of remaining choices. Specifically, it is easier to guess the third-ranked candidate than the fourth, despite having fewer options in the latter scenario. A plausible explanation is that many voters rank candidates from a single political party and then stop ranking others, and the more nuanced choice models are assigning high probability to candidates when other candidates in their political party are removed. For each dataset we show a histogram of the number of rankings of length $\ge k$. We see that there are many lists that rank the top few candidates and very few that rank all of the candidates, which means that the likelihood objective is more concerned with performing well near the top of the ranking than near the bottom. 

For the {\tt nascar} data, also in Figure~\ref{fig:logscr-plots}, we see that repeated selection with different choice models preforms roughly the same at the top positions, but PCMC outpaces the other models at predicting the lower finishers, particularly in the bottom 10 positions. The log probabilities exhibit much more variance across folds; there are far fewer races in a Winston Cup than ranked votes in the election datasets. The histogram shows us that nearly all of the races had between 34 and 37 racers finishing, of 76 total racers that finished at least one race in the cup. 

\begin{figure}[t]
\centering
\includegraphics[width=0.45\columnwidth]{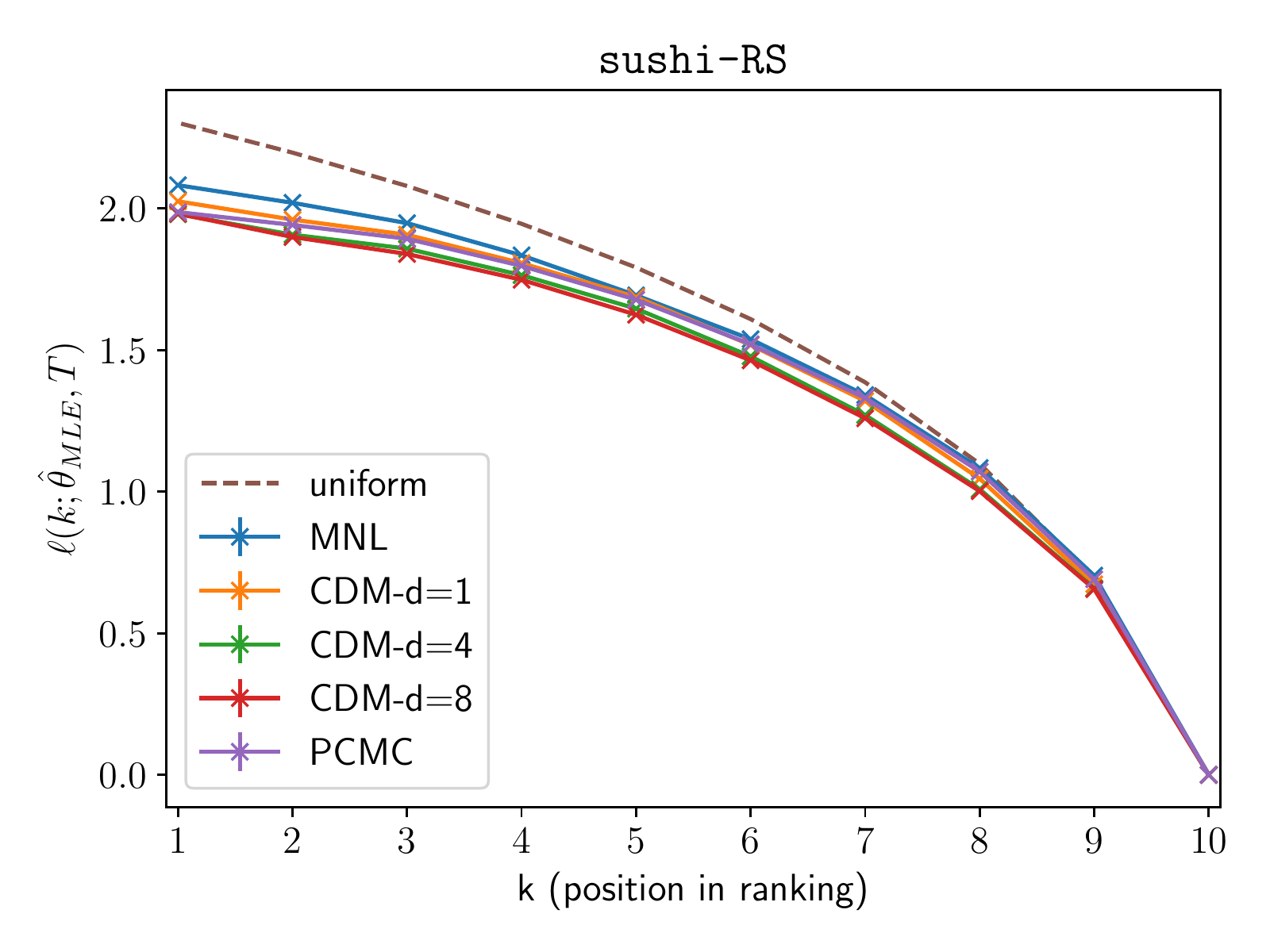}
\includegraphics[width=0.45\columnwidth]{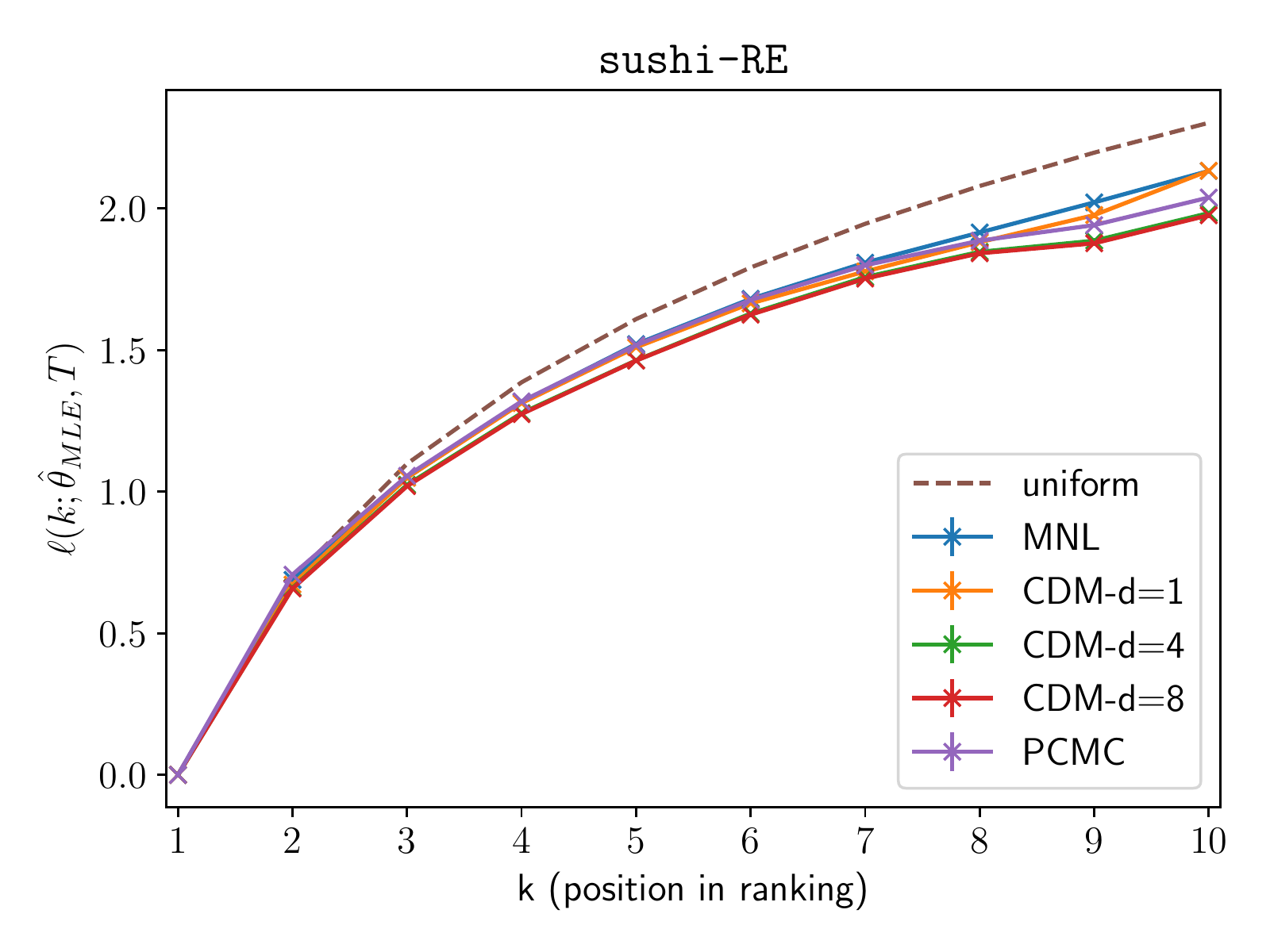}
\caption{The average position-level log likelihood of choice probabilities under repeated selection (left) and repeated elimination (right) on the {\tt sushi} dataset.}
\label{fig:sushi-plots}
\end{figure}

In Figure~\ref{fig:sushi-plots} we use the {\tt sushi} dataset, which contains complete rankings, to study the difference between $RS$ and $RE$ choice representations as a function of rank positions. For the $RE$-based models, $\sigma^{-1}(k)$ is chosen from $\{\sigma^{-1}(j)\}_{j\leq k}$, which has size $k$ rather than $n-k$ as with RS. Thus for $RS$ a uniform choice model assigns probability $\frac{1}{n-k+1}$ to position $k$, as there are $n-k+1$ unranked items when making the choice of position $k$, while for $RE$ the rankings are built from back to front and the uniform choice model assigns probability $\frac{1}{k}$ to each of the $k$ unranked items under consideration for position $k$. We can see that the last few positions in the $RE$-based models were slightly easier to predict than the first few positions under $RS$.

While we have been able to adapt repeated elimination to top-$k$ rankings, the choice set sizes are different for $RE$ and $RS$ for top-$k$ rankings and the heterogeneous normalization constants $Z_k(RE,p)= {n\choose k}$ are not easily attributed to position-specific choices. As a result, there is no straightforward way to compare position-level performance for $RE$ and $RS$ models unless the rankings are complete.


\section{Conclusions}

In this work we contribute a general framework for choice-based ranking by interpreting rankings as collections of choices, giving a method for translating any probabilistic discrete choice model into a probabilistic ranking model. We introduce the notion of a choice representation and focus our analysis on choice representations based on repeated selection and repeated elimination. We demonstrate that the only label-invariant choice representations that are normalized for prime $n$ are relabelings of repeated selection (which includes repeated elimination).

We extended repeated selection and repeated elimination to top-$k$ rankings and show that the normalization of the resulting distributions is independent of the choice model parameters for both repeated selection and repeated elimination. We showed that these normalization results allow us to train $RS$ and $RE$ models by applying standard estimation procedures to the choice data output by applying $RS$ and $RE$ to ranking data.

Because both the Mallows model and the Plackett-Luce model can be obtained through repeated selection, this choice-based framework serves as a simple conceptual tie between the extensive bodies of work that surround these two models. The framework further allows us to develop new ranking distributions by applying, e.g.,~repeated selection to the recently developed PCMC and CDM choice models. These models are not RUMs and can exhibit choice set effects (violations of the independence of irrelevant alternatives, IIA).

We examine the performance of several choice-based ranking models on a wide array of data including food preferences, elections, search engine rankings, and racing results, showing that repeated selection with the PCMC and CDM models outperform the seminal Plackett-Luce and (approximated) Mallows models on a wide variety of datasets. We thus find the $RS_{CDM}$ and $RS_{PCMC}$ to be tractable, flexible generalizations of the Plackett-Luce model with attractive structure and promising performance on a wide variety of datasets.

\xhdr{Acknowledgements}
This work was funded in part by a Dantzig-Lieberman Fellowship (SR), a Krishnan Shah Fellowship (SR), a David Morgenthaler II Faculty Fellowship (JU), and a Facebook Faculty Award (JU).

\bibliographystyle{apa}
\bibliography{ctrbibl}

\begin{thebibliography}{}

\bibitem[\protect\astroncite{Agarwal}{2016}]{agarwal2016ranking}
Agarwal, S. (2016).
\newblock On ranking and choice models.
\newblock In {\em IJCAI}, pages 4050--4053.

\bibitem[\protect\astroncite{Allison and Christakis}{1994}]{allison1994logit}
Allison, P.~D. and Christakis, N.~A. (1994).
\newblock Logit models for sets of ranked items.
\newblock {\em Sociological methodology}, pages 199--228.

\bibitem[\protect\astroncite{Azari~Soufiani
  et~al.}{2013a}]{soufiani2013generalized}
Azari~Soufiani, H., Chen, W., Parkes, D.~C., and Xia, L. (2013a).
\newblock Generalized method-of-moments for rank aggregation.
\newblock In {\em Advances in Neural Information Processing Systems}, pages
  2706--2714.

\bibitem[\protect\astroncite{Azari~Soufiani
  et~al.}{2013b}]{azari2013preference}
Azari~Soufiani, H., Parkes, D.~C., and Xia, L. (2013b).
\newblock Preference elicitation for general random utility models.
\newblock In {\em Uncertainty in Artificial Intelligence: Proceedings of the
  29th Conference}. AUAI Press.

\bibitem[\protect\astroncite{Bapat}{2003}]{bapat2003perron}
Bapat, R.~B. (2003).
\newblock Perron eigenvector of the {T}setlin matrix.
\newblock {\em Linear algebra and its applications}, 363:3--16.

\bibitem[\protect\astroncite{Ben-Akiva et~al.}{1985}]{ben1985discrete}
Ben-Akiva, M.~E., Lerman, S.~R., and Lerman, S.~R. (1985).
\newblock {\em Discrete choice analysis: theory and application to travel
  demand}, volume~9.
\newblock MIT press.

\bibitem[\protect\astroncite{Benson et~al.}{2016}]{benson2016relevance}
Benson, A.~R., Kumar, R., and Tomkins, A. (2016).
\newblock On the relevance of irrelevant alternatives.
\newblock In {\em Proceedings of the 25th International Conference on World
  Wide Web}, pages 963--973.

\bibitem[\protect\astroncite{Bhakta et~al.}{2013}]{bhakta2013mixing}
Bhakta, P., Miracle, S., Randall, D., and Streib, A.~P. (2013).
\newblock Mixing times of markov chains for self-organizing lists and biased
  permutations.
\newblock In {\em Proceedings of the twenty-fourth annual ACM-SIAM symposium on
  Discrete algorithms}, pages 1--15. SIAM.

\bibitem[\protect\astroncite{Blanchet et~al.}{2016}]{blanchet2016markov}
Blanchet, J., Gallego, G., and Goyal, V. (2016).
\newblock A markov chain approximation to choice modeling.
\newblock {\em Operations Research}, 64(4):886--905.

\bibitem[\protect\astroncite{Block and Marschak}{1960}]{block1960random}
Block, H.~D. and Marschak, J. (1960).
\newblock Random orderings and stochastic theories of responses.
\newblock {\em Contributions to Probability and Statistics}, 2:97--132.

\bibitem[\protect\astroncite{Boyd and Mellman}{1980}]{boyd1980effect}
Boyd, J.~H. and Mellman, R.~E. (1980).
\newblock The effect of fuel economy standards on the us automotive market: an
  hedonic demand analysis.
\newblock {\em Transportation Research Part A: General}, 14(5-6):367--378.

\bibitem[\protect\astroncite{Chen and Suh}{2015}]{chen2015spectral}
Chen, Y. and Suh, C. (2015).
\newblock Spectral mle: Top-k rank aggregation from pairwise comparisons.
\newblock In {\em International Conference on Machine Learning}, pages
  371--380.

\bibitem[\protect\astroncite{Critchlow et~al.}{1991}]{critchlow1991probability}
Critchlow, D.~E., Fligner, M.~A., and Verducci, J.~S. (1991).
\newblock Probability models on rankings.
\newblock {\em Journal of mathematical psychology}, 35(3):294--318.

\bibitem[\protect\astroncite{D{\'e}sir et~al.}{2016}]{desir2016assortment}
D{\'e}sir, A., Goyal, V., Jagabathula, S., and Segev, D. (2016).
\newblock Assortment optimization under the mallows model.
\newblock In {\em Advances in Neural Information Processing Systems}, pages
  4700--4708.

\bibitem[\protect\astroncite{Diaconis}{1988}]{diaconis1988group}
Diaconis, P. (1988).
\newblock Group representations in probability and statistics.
\newblock In {\em Lecture Notes-Monograph Series}. Institute for Mathematical
  Statistics.

\bibitem[\protect\astroncite{Dobrow and Fill}{1995}]{dobrow1995move}
Dobrow, R.~P. and Fill, J.~A. (1995).
\newblock The move-to-front rule for self-organizing lists with markov
  dependent requests.
\newblock In {\em Discrete Probability and Algorithms}, pages 57--80. Springer.

\bibitem[\protect\astroncite{Dwork et~al.}{2001}]{dwork2001rank}
Dwork, C., Kumar, R., Naor, M., and Sivakumar, D. (2001).
\newblock Rank aggregation methods for the web.
\newblock In {\em Proceedings of the 10th international conference on World
  Wide Web}, pages 613--622.

\bibitem[\protect\astroncite{Evans et~al.}{2014}]{evans2014leading}
Evans, S.~N., Rivest, R.~L., and Stark, P.~B. (2014).
\newblock Leading the field: Fortune favors the bold in {T}hurstonian choice
  models.
\newblock {\em arXiv preprint arXiv:1409.5924}.

\bibitem[\protect\astroncite{Farias et~al.}{2013}]{farias2013nonparametric}
Farias, V.~F., Jagabathula, S., and Shah, D. (2013).
\newblock A nonparametric approach to modeling choice with limited data.
\newblock {\em Management Science}, 59(2):305--322.

\bibitem[\protect\astroncite{Fligner and Verducci}{1986}]{fligner1986distance}
Fligner, M.~A. and Verducci, J.~S. (1986).
\newblock Distance based ranking models.
\newblock {\em Journal of the Royal Statistical Society. Series B
  (Methodological)}, pages 359--369.

\bibitem[\protect\astroncite{Hajek et~al.}{2014}]{hajek2014minimax}
Hajek, B., Oh, S., and Xu, J. (2014).
\newblock Minimax-optimal inference from partial rankings.
\newblock In {\em Advances in Neural Information Processing Systems}, pages
  1475--1483.

\bibitem[\protect\astroncite{Hausman and Ruud}{1987}]{hausman1987specifying}
Hausman, J.~A. and Ruud, P.~A. (1987).
\newblock Specifying and testing econometric models for rank-ordered data.
\newblock {\em Journal of econometrics}, 34(1-2):83--104.

\bibitem[\protect\astroncite{Kamishima}{2018}]{sushiwarning}
Kamishima, T. (2018).

\bibitem[\protect\astroncite{Kenyon-Mathieu and Schudy}{2007}]{kenyon2007rank}
Kenyon-Mathieu, C. and Schudy, W. (2007).
\newblock How to rank with few errors.
\newblock In {\em Proceedings of the thirty-ninth annual ACM symposium on
  Theory of computing}, pages 95--103. ACM.

\bibitem[\protect\astroncite{Khetan and Oh}{2016}]{khetan2016data}
Khetan, A. and Oh, S. (2016).
\newblock Data-driven rank breaking for efficient rank aggregation.
\newblock {\em Journal of Machine Learning Research}, 17(193):1--54.

\bibitem[\protect\astroncite{Khetan and Oh}{2018}]{khetan2018generalized}
Khetan, A. and Oh, S. (2018).
\newblock Generalized rank-breaking: computational and statistical tradeoffs.
\newblock {\em The Journal of Machine Learning Research}, 19(1):983--1024.

\bibitem[\protect\astroncite{Kingma and Ba}{2014}]{kingma2014adam}
Kingma, D.~P. and Ba, J. (2014).
\newblock Adam: A method for stochastic optimization.
\newblock {\em arXiv preprint arXiv:1412.6980}.

\bibitem[\protect\astroncite{Luce}{1959}]{luce1959individual}
Luce, R.~D. (1959).
\newblock {\em Individual Choice Behavior a Theoretical Analysis}.
\newblock John Wiley and sons.

\bibitem[\protect\astroncite{Luce}{1977}]{luce1977choice}
Luce, R.~D. (1977).
\newblock The choice axiom after twenty years.
\newblock {\em Journal of mathematical psychology}, 15(3):215--233.

\bibitem[\protect\astroncite{Mallows}{1957}]{mallows1957non}
Mallows, C.~L. (1957).
\newblock Non-null ranking models. i.
\newblock {\em Biometrika}, 44(1/2):114--130.

\bibitem[\protect\astroncite{Marden}{1996}]{marden1996analyzing}
Marden, J.~I. (1996).
\newblock {\em Analyzing and modeling rank data}.
\newblock CRC Press.

\bibitem[\protect\astroncite{Mattei and Walsh}{2013}]{MaWa13a}
Mattei, N. and Walsh, T. (2013).
\newblock Preflib: A library of preference data \textsc{http://preflib.org}.
\newblock In {\em Proceedings of the 3rd International Conference on
  Algorithmic Decision Theory (ADT 2013)}, Lecture Notes in Artificial
  Intelligence. Springer.

\bibitem[\protect\astroncite{McFadden}{1978}]{mcfadden1978modeling}
McFadden, D. (1978).
\newblock Modeling the choice of residential location.
\newblock {\em Transportation Research Record}, (673).

\bibitem[\protect\astroncite{McFadden}{1986}]{mcfadden1986choice}
McFadden, D. (1986).
\newblock The choice theory approach to market research.
\newblock {\em Marketing science}, 5(4):275--297.

\bibitem[\protect\astroncite{Negahban et~al.}{2016}]{negahban2016rank}
Negahban, S., Oh, S., and Shah, D. (2016).
\newblock Rank centrality: Ranking from pairwise comparisons.
\newblock {\em Operations Research}, 65(1):266--287.

\bibitem[\protect\astroncite{Negahban et~al.}{2018}]{negahban2018learning}
Negahban, S., Oh, S., Thekumparampil, K.~K., and Xu, J. (2018).
\newblock Learning from comparisons and choices.
\newblock {\em Journal of Machine Learning Research}, 19(40).

\bibitem[\protect\astroncite{Peysakhovich and Ugander}{2017}]{cdm}
Peysakhovich, A. and Ugander, J. (2017).
\newblock Learning context-dependent preferences from raw data.
\newblock In {\em Proceedings of the 12th workshop on the Economics of
  Networks, Systems and Computation}, page~8. ACM.

\bibitem[\protect\astroncite{Plackett}{1968}]{plackett1968random}
Plackett, R.~L. (1968).
\newblock Random permutations.
\newblock {\em Journal of the Royal Statistical Society. Series B}, pages
  517--534.

\bibitem[\protect\astroncite{Qin et~al.}{2010}]{qin2010new}
Qin, T., Geng, X., and Liu, T.-Y. (2010).
\newblock A new probabilistic model for rank aggregation.
\newblock In {\em Advances in Neural Information Processing Systems}, pages
  1948--1956.

\bibitem[\protect\astroncite{Ragain and Ugander}{2016}]{ragain2016pairwise}
Ragain, S. and Ugander, J. (2016).
\newblock Pairwise choice markov chains.
\newblock In {\em Advances in Neural Information Processing Systems}, pages
  3198--3206.

\bibitem[\protect\astroncite{Rajkumar and
  Agarwal}{2014}]{rajkumar2014statistical}
Rajkumar, A. and Agarwal, S. (2014).
\newblock A statistical convergence perspective of algorithms for rank
  aggregation from pairwise data.
\newblock In {\em ICML}, pages 118--126.

\bibitem[\protect\astroncite{Rivest}{1976}]{rivest1976self}
Rivest, R. (1976).
\newblock On self-organizing sequential search heuristics.
\newblock {\em Communications of the ACM}, 19(2):63--67.

\bibitem[\protect\astroncite{Sculley}{2010}]{sculley2010combined}
Sculley, D. (2010).
\newblock Combined regression and ranking.
\newblock In {\em Proceedings of the 16th ACM SIGKDD international conference
  on Knowledge discovery and data mining}, pages 979--988. ACM.

\bibitem[\protect\astroncite{Serfozo}{2009}]{serfozo2009basics}
Serfozo, R. (2009).
\newblock {\em Basics of applied stochastic processes}.
\newblock Springer Science \& Business Media.

\bibitem[\protect\astroncite{Talluri and van Ryzin}{2004}]{talluri2004revenue}
Talluri, K. and van Ryzin, G. (2004).
\newblock Revenue management under a general discrete choice model of consumer
  behavior.
\newblock {\em Management Science}, 50(1):15--33.

\bibitem[\protect\astroncite{Tversky}{1972}]{tversky1972elimination}
Tversky, A. (1972).
\newblock Elimination by aspects: A theory of choice.
\newblock {\em Psychological review}, 79(4):281.

\bibitem[\protect\astroncite{van Ryzin and Vulcano}{2017}]{van2017expectation}
van Ryzin, G. and Vulcano, G. (2017).
\newblock An expectation-maximization method to estimate a rank-based choice
  model of demand.
\newblock {\em Operations Research}, 65(2):396--407.

\bibitem[\protect\astroncite{Yellott}{1980}]{yellott1980generalized}
Yellott, J.~I. (1980).
\newblock Generalized thurstone models for ranking: equivalence and
  reversibility.
\newblock {\em Journal of Mathematical Psychology}, 22(1):48--69.

\bibitem[\protect\astroncite{Yue et~al.}{2007}]{yue2007support}
Yue, Y., Finley, T., Radlinski, F., and Joachims, T. (2007).
\newblock A support vector method for optimizing average precision.
\newblock In {\em Proceedings of the 30th annual international ACM SIGIR
  conference on Research and development in information retrieval}, pages
  271--278. ACM.

\end{thebibliography}

\renewcommand\thefigure{\thesection.\arabic{figure}}    
\setcounter{figure}{0}

\newpage

\begin{appendix}
\section*{Appendix}
This appendix begins by providing proofs that were omitted in the main paper. We then transition to additional expositions that may be useful to the interested reader. 
We provide an elaboration on the differences between repeated selection and repeated elimination, showing that these two choice representations yield strictly different sets of ranking models when applied to any family of choice models subsuming the MNL model family for as few as four alternatives. 
We then given an extended discussion of the connections between the present work and other Markov chain-based models for rankings, specifically Markov chains on $S_n$ itself. 
We conclude with a simple but useful derivation of the maximum likelihood estimate of the Mallows concentration parameter for partial rankings, conditional on a given reference permutation.

\section{Additional Proofs}
\label{app:proofs}
Here we present proofs that did not appear in the main paper. As in the main paper, we consider rankings $\sigma$ of $[n]$, an arbitrary indexing of the universe $U$ of alternatives. 

\subsection{Theorem~\ref{thm:Z=1revprime}}

We begin with the proof of Theorem~\ref{thm:Z=1revprime}, which shows that when ranking a prime number of alternatives, if $c_p$ distributions are already unit normalized for any choice model $p$, i.e. $Z(p,c)=1$ for any $p$, and $c$ is label-invariant, then $c$ must be $RS(\sigma \tau)$ for some $\tau \in S_n$. 

\begin{restatethm}[Restatement of Theorem~\ref{thm:Z=1revprime}]
Suppose $n$ is prime and $Z(p,c)$ for some label-invariant choice representation $c$ for all choice models $p$. Then $c(\sigma) = RS(\sigma \tau)$ for some $\tau \in S_n$. 
\end{restatethm}

In this proof we must prove that the choices in $c$ are simply relabelings of the choices in $RS$, but we only know that $c$ is label-invariant and that all $c_p$ distributions are unit normalized. In order to prove the theorem, we will show that considering certain families of choice models $p$ allow us to glean structure for $c$ from $Z(p,c)=1$. We use this technique several times and split its application into separate lemmas. 

As a summary of the proof technique, Lemma~\ref{lemma:bijection} tells us that the choices given by $c(\sigma)$ are ``in order" for some ranking $\tau$, and that this relationship is a bijection. Lemma~\ref{lemma:prodsetsize} tells us that the product of the choice set sizes is $n!$. Lemma~\ref{lemma:repeatedchoice} builds upon these two results to tell us that when an item is repeatedly chosen, the product of the corresponding choice set sizes is $n$. While these lemmas hold individually for all $n$, when $n$ is prime we are able to leverage Lemma~\ref{lemma:repeatedchoice} to argue that no item is repeatedly chosen and then apply the other lemmas to constrain the choice representation $c$ to have the desired structure ($c(\sigma) = RS(\sigma \tau)$, for some $\tau \in S_n$). The lemmas may also be of independent interest as they build intuition for the space of label-invariant choice representations.

Throughout the statements and proofs of the lemmas, we'll make extensive use of the shorthand
\begin{equation}
\sigma_{\min}(S) := \min_{j \in S} \sigma(j),
\end{equation}
which allows us to write $\sigma(i) = \sigma_{\min}(S)$ for alternative $i$ and subset $S$ to denote that under the ranking $\sigma$, $i$ is placed ahead of all of the alternatives in $S$. 

We now proceed to the first lemma, which shows that each ranking $\sigma \in S_n$ has its choices, $c(\sigma)$, ``in order" under precisely one $\tau \in S_n$, and that this relation gives a bijection. 

\begin{lemma}
If $Z(p,c)=1$ for label-invariant choice representation $c$ and any choice model $p$, then there exists a bijection $f:S_n \ra S_n$ such that $f(\sigma)=\tau$ if and only if $\sigma(i) = \sigma_{\min}(S)$ for all $(i,S) \in c(\tau)$. 
\label{lemma:bijection}
\end{lemma}
\proof
Here we consider ``deterministic" choice models $p_\sigma$ which always choose the alternative ranked first according to $\sigma$ among those in the choice set, i.e. $p_\sigma(i,S) = \1[\sigma(i) = \sigma_{\min}(S)]$. Because $p_{\sigma}(i,S) \in \{0,1\}$ for every alternative $i$ and subset $S$, the product of any choice probabilities under $p_\sigma$ is also 0 or 1. It follows that for any permutation $\tau \in S_n$, $\prod_{(i,S) \in c(\tau)} p_{\sigma}(i,S) \in \{0,1\}$. 

It follows from $Z(p_\sigma,c) = 1$ that $\sum_\tau  \prod_{(i,S) \in c(\tau)} p_{\sigma}(i,S) =1$, so there must be exactly one $\tau$ in $S_n$ such that the product of choice probabilities with $p_\sigma$ over $c(\tau)$ is 1, proving that $f(\sigma) = \tau$ is well-defined. Because $f$ is from $S_n$ to itself, we now need only show that $f$ is one-to-one to show that it is a bijection. 

To show that $f$ is one-to-one, we leverage label-invariance. Suppose $f(\sigma) = \tau$. For any $x \in S_n$, let $x^{-1}$ denote the unique inverse of $x$. By the definition of label-invariance (Definition~\ref{def:li}), 
\[
c(\tau x) = \{(x^{-1}(i),x^{-1}(S)) : (i,S) \in c(\tau)\}.
\]

Consider now the choice model $p_{\sigma x}$. 
Because $\sigma(x(x^{-1}(j))) = \sigma(j)$ for all $j$, we have that $\min_{j \in S} [\sigma xx^{-1}(j)] = \sigma x x^{-1}_{\min}(S) = \sigma_{\min}(S)$ and thus 
\begin{align*}
p_{\sigma x}(x^{-1}(i),x^{-1}(S)) &= \1\left[\sigma(x(x^{-1}(i))) = \sigma x x^{-1}_{\min}(S) \right] \\
&= \1\left[ \sigma(i) = \sigma_{\min}(S)\right]\\
&= p_{\sigma}(i,S).
\end{align*}

It follows that we have that $f(\sigma) = \tau$ if and only if $f(\sigma x) = \tau x$ for all $x \in S_n$. 

Now let $e \in S_n$ denote the identity permutation, i.e.~$e(i) = i$ for $i \in [n]$. Let $f(e) = z$. Then $f(x) = f(ex)=zx$ for any $x \in S_n$. Because composition with $z$ is an automorphism on $S_n$, if $x \neq y$, $f(x) = zx \neq zy = f(y)$. We conclude that $f$ is a bijection. 
\endproof

\begin{lemma}
Suppose $Z(p,c)=1$ for label-invariant choice representation $c$ and any choice model $p.$  For any $\sigma \in S_n$, let $c(\sigma) = \{(i_k,S_k)\}_{k=1}^m$. Then 
\begin{equation}
\prod_{k=1} |S_k| = n! 
\end{equation}
\label{lemma:prodsetsize}
\end{lemma}

\proof
To begin, note that the result is immediate for $c=RS$. Indexing $(i,S) \in RS(\sigma)$ so that $i_k = \sigma^{-1}(k)$ and $S_k = \{\sigma^{-1}(j)\}_{j \geq k}$ for $k \in [n-1]$:
\begin{align*}
\prod_{(i_k,S_k) \in RS(\sigma)}|S_k| &= \prod_k^{n-1} | \{\sigma^{-1}(j)\}_{j \geq k}|\\
&= \prod_{k=1}^{n-1} n-k+1 \\
&= n!.
\end{align*}
To show that it holds for any label-invariant choice representation $c$ with $Z(p,c)=1$ for any $p$, we will again consider a specific choice model, this time $p_{unif}$, which chooses uniformly from any choice set, and show that it gives the desired result. 

We begin by noting that by label-invariance, the choice sets in $c(\sigma)$ have the same sizes for every sigma. 
Consider the choice model $p_{unif}(i,S) = \frac{1}{|S|}$. We have that for every $\sigma$, 
\[
\prod_{(i,S) \in c(\sigma) } p_{unif}(i,S) = \prod_{k=1}^m \frac{1}{|S_k|}.
\]
It follows that 
\begin{align*}
Z(p_{unif},c) &= \sum_\sigma \prod_{(i,S) \in c(\sigma) } p_{unif}(i,S) \\
&= \sum_{\sigma}  \prod_{k=1}^m \frac{1}{|S_k|} \\
&= \frac{n!}{\prod_{k=1}^m |S_k|}.
\end{align*}
Where the last equality comes from noting that we are adding the same term for all $n!$ rankings. 
Now because $Z(p_{unif},c) =1$ we have that
$\prod_{k=1}^m |S_k|=n!$, 
the desired result. 
\endproof

Our final lemma will show that when an alternative is chosen multiple times, we can leverage a choice model that chooses that alternative whenever it is present to prove that the product of the sizes of the choice sets from which that alternative is selected is $n$. 

\begin{lemma}
Let $Z(p,c)=1$ for all choice models $p$ for label-invariant $c$. Suppose there exists an alternative $j$ such that $(j,S_1),(j,S_2), \dots, (j,S_\ell)$ are all in $c(\tau)$ for some $\tau$. Then $\prod_{k=1}^\ell |S_k| = n$. 
\label{lemma:repeatedchoice}
\end{lemma}

\proof
For each alternative $j \in [n]$, we consider a choice model $p_j$ which chooses $j$ if it is in the choice set, and chooses uniformly from the choice set when $j$ is not present, i.e. $p_j(i,S) = \1(j = i)\1(j \in S) + 1(j \not \in S)\frac{1}{|S|}$. 

Recall that from Lemma~\ref{lemma:bijection} that there exists bijection $f$ with $f(\sigma) = \tau$ if $\sigma(i) = \sigma_{\min}(S)$ for every $(i,S) \in c(\tau)$. Because $f$ is a bijection, it is invertible- for every $\tau \in S_n$, there exists a unique $\sigma= f^{-1}(\tau)$ such that $\sigma(i) = \sigma_{\min}(S)$ for all $(i,S) \in c(\tau)$. Note that $\prod_{(i,S) \in c(\tau)} p_j(i,S) > 0$ only if $j=i$ for all $(i,S) \in c(\tau)$ with $S \ni j$. Because $f^{-1}(\tau)(i) = f^{-1}(\tau)_{\min}(S)$ for all $(i,S) \in c(\tau)$, it follows that
\begin{equation}
\prod_{(i,S) \in c(\tau)} p_j(i,S) > 0 \Leftrightarrow f^{-1}(\tau)(j) = 1. 
\end{equation}
In other words, $j$ must be ranked  ``in first place" by $f^{-1}(\tau)$ for the product of choice probabilities with $p_j$ over $c(\sigma)$ to be positive. 

Suppose $f^{-1}(\tau)(j)=1$, and index the choices $c(\tau) = \{(i_k,S_k)\}_{k=1}^m$ so that the first $\ell$ choices are $(j,S_1),\dots,(j,S_\ell)$. Now $p_j(i_k,S_k) = 1$ if $k\leq \ell$ and $1/|S_k|$ otherwise. It follows from Lemma \ref{lemma:prodsetsize} that $\prod_{k=1}^m |S_k| = n!$, so for every $\tau$ such that $f^{-1}(\tau)(j)=1$:
\begin{align*}
\prod_{(i,S) \in c(\tau) : f^{-1}(\tau)(j) = 1} p_j(i,S) &= \prod_{k=1}^m p_j(i_k,S_k) \\
&= \prod_{k = \ell+1}^m \frac{1}{|S_k|} \\ 
&=\frac{\prod_{k\leq \ell} |S_k|}{\prod_{k=1}^m |S_k|} \\
&=\frac{\prod_{k=1}^\ell |S_k|}{n!}.
\end{align*}
We now note that there that there are $(n-1)!$ rankings $\sigma$ with $\sigma(j) = 1$, so there are $(n-1)!$ rankings $\tau$ such that $f^{-1}(\tau)(j)=1$. It follows from $Z(p_j,c) = 1$ that
\begin{align*}
\sum_\tau \prod_{(i,S) \in c(\tau)} p_j(i, S) &=\sum_{\sigma: \sigma(j) = 1} \frac{\prod_{k = 1}^\ell |S_k|}{n!} \\
&= (n-1)! \frac{\prod_{k = 1}^\ell |S_k|}{n!} \\
&= \frac{\prod_{k = 1}^\ell |S_k|}{n} = 1.
\end{align*}
Multiplication by $n$ thus gives the desired result.
\endproof

With these lemmas in place, we now turn to proof of Theorem~\ref{thm:Z=1revprime}. We'll leverage Lemma 1 to argue that $c(\sigma)$ contains choices where $\tau^{-1}(k)$ is chosen from a choice set containing $\tau^{-1}(k+1)$ for every $k$ where $\tau=f(\sigma)$. Then we'll use Lemma 3 to argue that there are exactly $n-1$ choices in $c(\tau)$, and argue that to satisfy Lemma 2, $\tau^{-1}(k)$ must be chosen from the largest possible set, namely $\{\tau^{-1}(j)\}_{j \geq k}$, which will give us $c(\sigma) = RS(\sigma \tau^{-1})$. 

\vspace{5mm}

\proof ({\it Theorem~2})

Let $n$ be a prime and let $Z(p,c) = 1$ for a label-invariant choice representation $c$ and all choice models $p$ on $U = [n]$. Let $f$ be the bijection given by Lemma~\ref{lemma:bijection}, and let $f(\sigma) = \tau$, so $\sigma$ is the unique ranking with $\sigma(i) = \sigma_{\min}(S)$ for each $(i,S) \in c(\tau)$. By the uniqueness of $\sigma$, it follows that for every $\ell \in [n]$, there exists some $(\sigma^{-1}(\ell),S)\in c(\tau)$ with $S \ni \sigma^{-1}(\ell+1)$. otherwise we could swap $\sigma^{-1}(\ell)$ and $\sigma^{-1}(\ell+1)$ while still maintaining the ordering of the choices in $c(\tau)$. This guarantees at least $n-1$ choices in $c(\tau)$. 

Noting that $\sigma^{-1}(n)$ cannot be chosen from any sets, as we do not allow choice sets to be singletons, we now claim there are exactly $n-1$ choice sets in $c(\tau)$. We proceed by contradiction. Suppose there were $n$ or more choices in $c(\tau)$. Then, by the pigeonhole principle, there would be some item $i$ chosen twice, and by Lemma~\ref{lemma:repeatedchoice}, the product of the choice set sizes would be $n$. However, choice set sizes are at least 2 and $n$ is prime, so this is impossible. It follows that there are exactly $n-1$ choice sets in $c(\tau)$, and each item with $\sigma(i)<n$ is chosen exactly once. 

Because $\sigma(i) = \sigma_{\min}(S)$ for each $(i,S) \in c(\tau)$, it follows from $\sigma(i)=k$ that $\sigma(j) \geq k$ for all $j \in S$. Therefore, $\{\sigma(j) : j \in S\} \subseteq\{k,k+1,\dots,n\}$, and thus $|\{\sigma(j) : j \in S\}| \leq n-k+1$. Index the choices in $c(\tau)$ by $\sigma(i)$: $c(\tau) = \{(\sigma^{-1}(k), S_k)\}_{k=1}^{n-1}$. We have from Lemma \ref{lemma:prodsetsize} that $\prod_{k=1}^{n-1} |S_k| = n!$ and also that $|S_k| \leq n-k+1$. Because $\prod_{k=1}^{n-1} (n-k+1) = n!$, each choice set must achieve its upper bound for size, and thus 
\[
c(\tau) = \{(\sigma^{-1}(k),\{\sigma^{-1}(j)\}_{j\geq k})\}_{k=1}^{n-1} = RS(\tau\sigma).
\]
Note that the last equality is the definition of label-invariance. 
\endproof

With the proof of Theorem~2 complete, we note that the primality of $n$ was only used to eliminate the appearance of repeated choices of the same alternative, and that we have no counterexamples of the theorem for composite $n$. Note also that the general method in the proof, constructing choice models $p$ such that $Z(p,c)=1$ gives us structure on label-invariant $p$, was not exhausted. We conjecture that Theorem~\ref{thm:Z=1revprime} can be extended to composite $n$ with one or more additional instructive choices of $p$. 

\begin{conjecture}
Theorem~\ref{thm:Z=1revprime} can be extended to all integers $n \geq 2$. 
\end{conjecture}

\subsection{Proof of Theorem~\ref{thm:ZkRE=nchoosek}}

\begin{restatethm}[Restatement of Theorem~\ref{thm:ZkRE=nchoosek}]
For $k \in \{2,\dots,n\}$, $Z_k(RE,p) = {n \choose k}$ for any choice model $p$. 
\end{restatethm}

\proof 
We proceed by induction, down from $n$ to $k$. For the base case, we simply note that top-$n$ rankings are full rankings and we have that $Z_{n}(RE,p) = Z(RE,p) = 1 = {n \choose n}$, by Theorem~\ref{thm:Z=1}. 

Now we assume the result for $\ell \in \{ k+1,\dots,n\}$. Let $q$ be any choice model, and let $p$ be a choice model that matches $q$ on all sets except possibly for those of size $k+1$. Specifically, we have $p(i,S) = 1/(k+1)$ whenever $|S| = k+1$, and $p(i,S) = q(i,S)$ otherwise. Now $p$ and $q$ agree on all sets of size $\leq k$, so by Observation~\ref{obs:retopksetsize}, $Z_k(RE,p) = Z_k(RE,q)$. 

Manipulating the definition of $Z_{k+1}(RE,p)$, we can move out the choice probabilities for the sets of size $k+1$ as follows:
\begin{align*}
Z_{k+1}(RE,p) &= \sum_{\tau \in S_{k+1,n}} \prod_{(i,S) \in RE(\tau)} p(i,S) \\
&= \sum_{\tau \in S_{k+1,n}} \prod_{i=2}^{k+1} p(\tau^{-1}(i), \{\tau^{-1}(j)\}_{j\leq i})\\
&= \sum_{\tau \in S_{k+1,n}} p(\tau^{-1}(k+1), \{\tau^{-1}(j)\}_{j\leq i}\prod_{i=2}^{k} p(\tau^{-1}(i), \{\tau^{-1}(j)\}_{j\leq i})\\
&= \frac{1}{k+1}\sum_{\tau \in S_{k+1,n}} \prod_{i=2}^{k} p(\tau^{-1}(i), \{\tau^{-1}(j)\}_{j\leq i}).
\end{align*}

Partition $S_{k+1,n}$ into sets $A_\sigma$ indexed by lists $\sigma \in S_{k,n}$: $A_\sigma = \{ \tau \in S_{k+1,n} : \tau(i) = \sigma(i), \forall i \in [k] \}$. This partitioning divides $S_{k+1,n}$ according to the prefix alignment with $S_{k,n}$.
 Note that $|A_\sigma| = n-k$ for every $\sigma$, as we can append any of the $n-k$ items not ranked by a top-$k$ list to obtain a top-$(k+1)$ list. If $\tau \in A_{\sigma}$, then $\prod_{i=2}^{k} p(\tau^{-1}(i), \{\tau^{-1}(j)\}_{j\leq i}) = \prod_{(i,S) \in RE(\sigma)} p(i,S)$. 
 
 Substituting this expression into our equation for $Z_{k+1}(RE,p)$ above gives: 
 \begin{align*}
  Z_{k+1}(RE,p)&=\frac{1}{k+1}\sum_{\tau \in S_{k+1,n}} \prod_{i=2}^{k} p(\tau^{-1}(i), \{\tau^{-1}(j)\}_{j\leq i}) \\
  &=   \frac{1}{k+1}\sum_{\tau \in A_\sigma} \prod_{i=2}^{k} p(\tau^{-1}(i), \{\tau^{-1}(j)\}_{j\leq i}) \\
&=   \frac{1}{k+1}\sum_{\tau \in A_\sigma}\prod_{(i,S) \in RE(\sigma)} p(i,S) \\
 &= \frac{1}{k+1}\sum_{\sigma \in S_{k,n}} |A_\sigma|  \prod_{(i,S) \in RE(\sigma)} p(i,S) \\
 &= \frac{n-k}{k+1} \sum_{\sigma \in S_{k,n}} \prod_{(i,S) \in RE(\sigma)} p(i,S)\\
 &= \frac{n-k}{k+1} Z_k(RE,p).
 \end{align*}
 Recalling that $Z_{n}(RE,p) = 1$ and assuming the inductive hypothesis, we thus have 
 \[
 Z_k(RE,p) = \frac{k+1}{n-k} \cdot Z_{k+1}(RE,p) 
 = \frac{k+1}{n-k} \cdot {n \choose k+1} 
 = {n \choose k}.
 \]
Since $Z_k(RE,q)=Z_k(RE,p)$ for all choice models $q$ by Observation~\ref{obs:retopksetsize}, the proof is complete.
\endproof 

\subsection{Other omitted proofs}

\begin{restateprop}[Restatement of Proposition~\ref{prop:PCMC-nest}]
For a PCMC choice model $p$ parameterized by rate matrix $Q$ and a partition $S_1,\dots,S_k$ of $U$ where for all $i \in S_I, j \in S_J$ with $I \neq J$, $q_{ij} = \lambda_{IJ}$, then for each $I$, $(p,S_I,U)$ exhibits nested IIA.
\end{restateprop}

\proof
The conditions given on $Q$ make $S_1,\dots,S_k$ meet the criteria for a contractible partition (Definition 3, \citep{ragain2016pairwise}), so by Proposition 3 of \citep{ragain2016pairwise}, for any subset $T$ of $U$, $p(i,T)=p(i,S_I \cap T)p(S_I \cap T, T)$ where  $p(i,S_I \cap T)$ are the choice probabilities given by the restriction of $Q$ to $S_I \cap T$ and $p(S_I \cap T,T)$ follows a separate PCMC choice model on the $S_1,\dots,S_k$ whose rate matrix is a function of the sizes of $S_I \cap T$ and the $\lambda_{IJ}$ giving the block structure.
\endproof

\begin{restateprop}[Restatement of Proposition~\ref{prop:CDM-nest}]
For a CDM choice model $p$ parameterized by pairwise contextual utilities $\mathcal{U}=\mathbf{f}^T\mathbf{c}$ with $\mathcal{U}_{ij}$ fixed for all $i \in S, j \not \in S$, $(p,S,U)$ exhibits nested IIA.
\end{restateprop}

\proof
Let $T$ be any subset of $U$ and let $C = (|T|-|S|)c$. For any $i \in S$ we have that $\sum_{k \in T} u_{ik} = C + \sum_{k \in S} u_{ik}$, and thus that $
\sum_{j \in S} e^{\sum_{k \in T} u_{jk}} = \sum_{j \in S} e^{\sum_{k \in S} u_{jk}}e^{C}.$

From these we have that 
\begin{align*}
p(i,S) &= \frac{e^{\sum_{k \in T} u_{ik}}}{\sum_{j \in T} e^{\sum_{k \in T} u_{ik}}}= \frac{e^{\sum_{k \in S} u_ik}e^C}{\sum_{j \in T} e^{\sum_{k \in T} u_{ik}}}\cdot \frac{\sum_{j \in S} e^{\sum_{k \in S} u_{jk}}}{\sum_{j \in S} e^{\sum_{k \in S} u_{jk}}} \\
&=\frac{\sum_{j \in S} e^{\sum_{k \in T} u_{jk}}}{\sum_{j \in T} e^{\sum_{k \in T} u_{ik}}}\cdot \frac{e^{\sum_{k \in S} u_ik}}{\sum_{j \in S} e^{\sum_{k \in S} u_{jk}}} = p(S,T)p(i,S).
\end{align*}
Note that we moved the $e^C$ term from the sum of $i$'s utility from $T-S$ into the sum for each element in $S$, which all have the same utility from $T-S$ as $i$ by the assumption.
\endproof

\begin{restateprop}[Restatement of Proposition~\ref{prop:rs-nested-IIA}]
For a choice model $p$ on universe $U$, suppose $(p,S,U)$ exhibits nested IIA. For $\sigma$ drawn from $P_{RS,p}$, let $\sigma_S$ be the restriction of $\sigma$ to $S$ and $\tilde p_S$ be the restriction of $p$ to $S$. Then $\sigma_S$ is distributed according to $P_{RS,\tilde p_S}$.
\end{restateprop}

\proof
Let $|S| = m$. We partition $S_n$ into ${n \choose m}$ subsets $X_v$ indexed by binary vectors $v$ of length $n$ with $\sigma \in X_v \Leftrightarrow v(i) = 1(\sigma^{-1}(i) \in S)$, i.e. the $m$ entries of $v$ with a one are the positions at which rankings in $X_v$ assign items of $S$. 

Then for $i \in S$ and any subset $T$ of $U$, we have by nested IIA that 
$$p(i,T) = p(i,S \cap T) p(S \cap T, T),$$ 
and if we condition on our choice from $T$ lying in $S$, the second term becomes 1 while the first is unchanged. 

This thus gives for any $v$, the distribution of $\sigma_S$ only depends on the choices made at the entries where alternatives in $S$ are ranked according to $v$, and the nested IIA allows us to turn these choices probabilities into those from the subsets of $S$ cleanly. Thus we have
\begin{align*}
Pr(\sigma_S | \sigma \in X_v) &= \prod_{\ell =1}^n v(\ell)p(\sigma^{-1}(\ell), \{\sigma^{-1}(j)\}_{j \geq \ell}) =  \prod_{\ell =1}^n v(\ell)p(\sigma^{-1}(\ell), S \cap\{\sigma^{-1}(j)\}_{j \geq \ell})\\
&= \prod_{k=1}^m p(\sigma_S^{-1}(k),\{\sigma_S(j)\}_{j \geq k})= P_{RS,p}(\sigma_S).
\end{align*}
Note that these steps did not depend on $v$, so the proposition follows. 
\endproof

The proposition gives a natural analog of nested IIA for rankings in that just as choices within a nest exhibit independence from irrelevant alternatives outside the nest, repeated selection with a choice model exhibiting nested IIA yields independence between the relative orderings of each of the nests.

\section{Repeated elimination and reversibility}
\label{app:reverse}

The natural counterpart to repeated selection ($RS$) is repeated elimination ($RE$) (as introduced in Section~\ref{sec:rsre}) and rankings models built up using $RS$ and $RE$ applied to the same choice model will intuitively differ quite dramatically for almost any choice model $p$. Less obviously, it's not clear how similar we can make the ranking distribution induced under $RS$ with choice model $p$ to the ranking distribution induced under RE with some choice model $p'$ from the same choice model family. In this section we show that even for simple families of choice models, most $RS$ ranking models will have no $RE$ counterpart that provides the same ranking distribution.

Recall that repeated elimination models the construction of a ranking $\sigma$ as the choice of the last item, $\sigma^{-1}(n)$, from $U$, then the second to last item, $\sigma^{-1}(n-1)$, as chosen from $U-\sigma^{-1}(n)$, and so forth. The intuition behind this representation is that the agent building the ranking eliminates the worst alternative, placing it at the back of $\sigma$, and then proceeds by eliminating the worst remaining alternative from what remains repeatedly until only $\sigma^{-1}(1)$ is left to place at the front of the ranking $\sigma$. 

We will show that choice-based ranking with repeated elimination can produce different classes of distributions than repeated selection, even when the choice models come from a simple MNL family over $n=3$ items. In concrete terms, the family of Plackett-Luce ranking distributions is different from the family of Plackett-Luce ranking distributions ``backwards.''

Critchlow et al.~give a general definition of reversibility that is a property of a family of ranking distributions and fits well into our framework. We will use this definition to define reversibility as a property of a family of choice models.

\begin{definition}[Reversibility of a family of ranking models, \citep{critchlow1991probability}]
Let $\pi$ be the permutation that reverses a ranking in the sense that $\pi(i) = n-i+1$. A family $\mathcal{P}$ of distributions on $S_n$ is reversible if for every distribution $P \in \mathcal{P}$ there exists $P' \in \mathcal{P}$ such that for all $\sigma$, $P(\sigma) = P'(\pi \sigma)$. 
\end{definition}

Critchlow et al.~showed that $L$-decomposable distributions on rankings
are reversible only if they are also $R$-decomposable, where $L$-decomposability is to repeated selection as $R$-decomposability is to repeated elimination. A distribution $D$ on $S_n$ is $R$-decomposable if and only if 
$$
Pr_D (\sigma(i)=x | \sigma^{-1}(k)=x_k, k>i) = Pr_D \left(\sigma(i)=x| \{\sigma^{-1}(k)\}_{k>i}\right), \forall \sigma \in S_n.
$$
We use this characterization of reversibility to extend the property to choice models, so that we can describe a family of choice models by whether or not the set of ranking distributions given by repeated selection with choice models in the family is the same as the set of ranking distributions given by repeated elimination with choice models in the family. 

\begin{definition}[Reversibility for choice models]
A family of choice models $\mathcal{P}$ is {\it reversible} if for every $p \in \mathcal{P}$ there exists some $p' \in \mathcal{P}$ such that $P_{RS,p}(\sigma) = P_{RE,p'}(\sigma)$ for all $\sigma \in S_n$.
\end{definition}

Critchlow et al.~showed that the Mallows ranking model is reversible, and it follows that the corresponding Mallows choice model exhibits reversibility. Specifically, for a Mallows choice model $p_{\sigma_0,\theta}$ as defined in Section~\ref{sec:rs}, the Mallows choice model $p'$ for which repeated elimination with $p'$ matches repeated selection with $p_{\sigma_0,\theta}$ has the same concentration parameter $\theta$ and the reversal of $\sigma_0$ as its reference permutation.

Meanwhile, we find that none of the other choice models we consider are reversible, with $RS$ and $RE$ giving different families of ranking models for these choice models. The absence of reversibility in CDM and PCMC stems from the lack of reversibility for the MNL model, a special case of each. To prove this rigorously we will use the following proposition due to Luce, which is simple but widely considered to be surprising \citep{yellott1980generalized,block1960random,luce1959individual}.

\begin{proposition}[MNL is not reversible for 3 or more items  \citep{luce1959individual}]
\label{prop:mnlnorev}
Repeated selection with MNL and repeated elimination with MNL yield the same distribution for rankings of at least three items only when both distributions are uniform.
\end{proposition}
\noindent

\begin{figure}
\centering
\includegraphics[scale=.4]{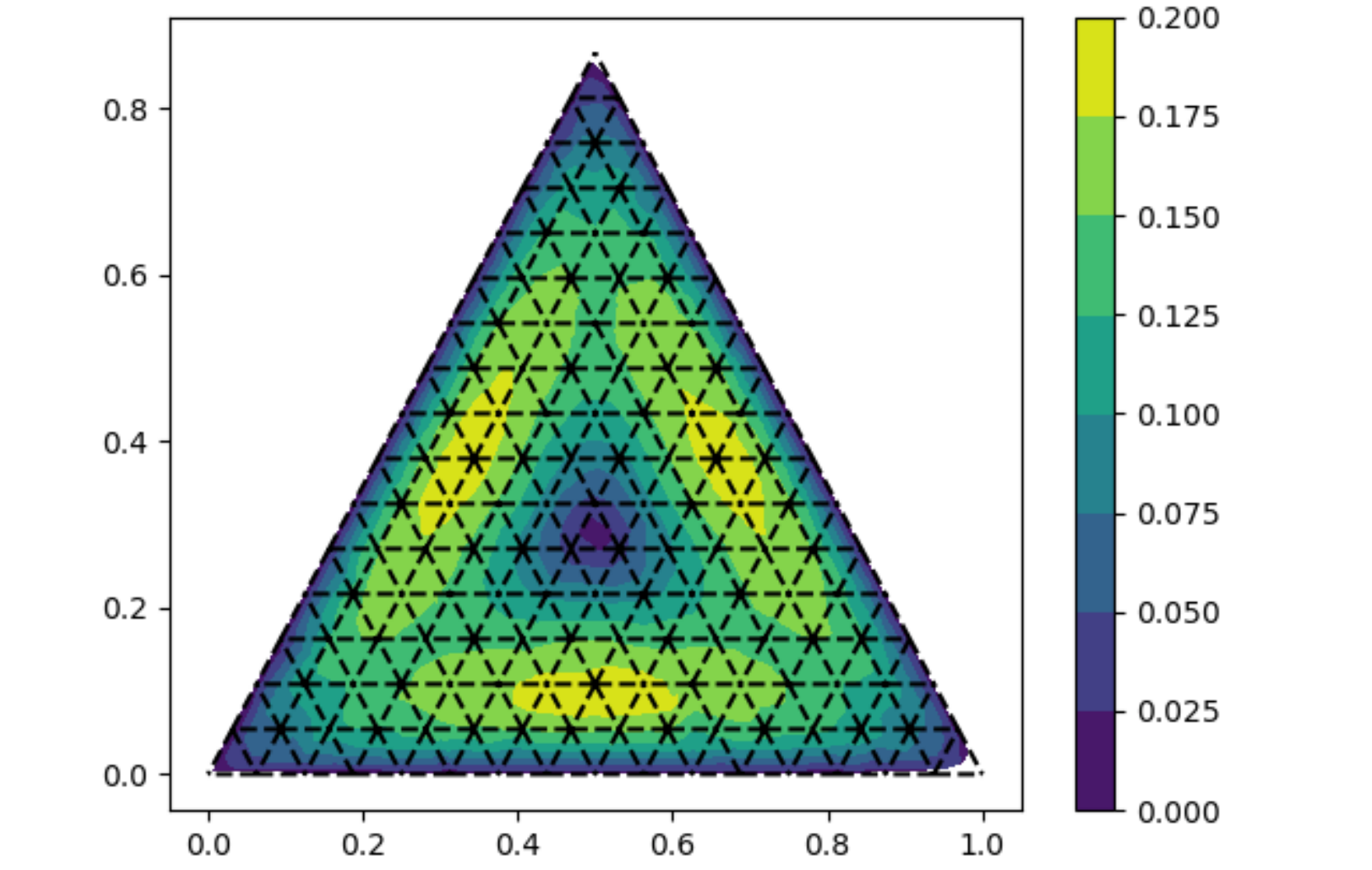}
\caption{The total variation distance from the $RS_{MNL}$ distribution (Plackett-Luce) on three items with parameters $(\gamma_1, \gamma_2, \gamma_3)$ and the nearest $RE_{MNL}$ distribution, plotted over the $\gamma$ 3-simplex. We see that the two models are equivalent in the center (the uniform distribution) and in the corners of the simplex.}
\label{fig:simplex}
\end{figure}

The notion of reversibility has also been applied to joint selection from RUM distributions. Yellott defines a family of RUMs as reversible if the ranking distributions that arises from sorting the random utilities in increasing order being the same as the family that arises from sorting the random utilities in decreasing order \citep{yellott1980generalized}. Yellott notes in particular that Plackett-Luce ranking models are not reversible unless all of the random utilities have the same location parameter (the same mean). Recalling that joint selection is equivalent to repeated selection with an MNL choice model \citep{block1960random}, this provides an alternative proof that $RS_{MNL}$ and $RE_{MNL}$ are different distributions (Proposition ~\ref{prop:mnlnorev}). 

When ranking three items, the family of $RS_{MNL}$ ranking models is parameterized by a 3-simplex of normalized MNL parameters $(\gamma_1, \gamma_2, \gamma_3)$. As a concrete demonstration, Figure~\ref{fig:simplex} illustrates the minimum total variation distance between a $RS_{MNL}$ model and the nearest $RE_{MNL}$ model over this simplex,
$$
\argmin_{\gamma'} d_{TV}(P_{RS,\gamma},P_{RE,\gamma'}),
$$
where $d_{TV}(p,q)$ is the total variation distance between distributions $p$ and $q$ on $S_n$. 
The minimum total variation distance between a given $RS_{MNL}$ model and the nearest $RE_{MNL}$ model can clearly be non-trivially greater than zero. Further exploration of this difference, including minimax bounds, would be an interesting direction for future research.

We can leverage the non-reversibility of MNL to show that models that include MNL as a special case are not reversible for $n\geq 4$ alternatives. We accomplish this by constructing a specific MNL model on four items whose repeated selection ranking distribution is not $R$-decomposable, and thus cannot be the repeated elimination distribution of any choice model.

\begin{proposition}
\label{prop:including-mnl-not-rev}
Any family of choice models on $n \ge 4$ alternatives that include all MNL models on $n$ alternatives is not reversible. 
\end{proposition}

\proof
We first give a simple counterexample to reversibility on four items. We then use show that this counterexample applies to all sets with $n\ge 4$ alternatives. The central idea is that $R$-decomposability constrains the ratio of probabilities of two similar lists in a way that cannot be captured by any choice model when those ranking probabilities come from certain Plackett-Luce models. 

Suppose there are exactly four alternatives, let $U = \{1,2,3,4\}$ and let $\gamma = (\gamma_1,\gamma_2,\gamma_3,\gamma_4)$ be the parameters of an MNL model. We consider the repeated selection distribution on $S_4$ with $\gamma$, and examine the probability of a fixed $\pi=\pi^{-1}(1)\pi^{-1}(2)\pi^{-1}(3)\pi^{-1}(4)=1243$ as well as the probability of the identity permutation $e=1234$. 
\begin{eqnarray*}
P_{RS,\gamma}(\pi) = \frac{\gamma_1}{\gamma_1+\gamma_2+\gamma_3+\gamma_4}\cdot \frac{\gamma_{2}}{\gamma_2+\gamma_3+\gamma_4}\cdot \frac{\gamma_4}{\gamma_3+\gamma_4}.
\\
P_{RS,\gamma}(e) = \frac{\gamma_1}{\gamma_1+\gamma_2+\gamma_3+\gamma_4}\cdot \frac{\gamma_{2}}{\gamma_2+\gamma_3+\gamma_4}\cdot \frac{\gamma_3}{\gamma_3+\gamma_4}.
\end{eqnarray*}
Noting that $P_{RS,\gamma}(\pi) = P_{RS,\gamma}(e) \Leftrightarrow \gamma_3=\gamma_4$, we thus have for $\gamma_3 \neq \gamma_4$ that  
\begin{align*}
P_{RS,\gamma}(\sigma(2)=2 | \sigma(3)=4,\sigma(4)=3) = P_{RS,\gamma}(\pi)
 \neq P_{RS,\gamma}(e) = P_{RS,\gamma}(\sigma(2)=2 | \sigma(3)=3,\sigma(4)=4).
 \end{align*}
It follows that $P_{RS,\gamma}$ is not $R$-decomposable, so it cannot be the repeated elimination distribution of any choice model $p$.

If $n$, the number of alternatives, is greater than 4, we can simply let $\tilde \gamma = (\gamma_1-\epsilon,\gamma_2-\epsilon,.\gamma_3-\epsilon,\gamma_4-\epsilon,\delta,\delta^2,\dots,\delta^{n-4})$ where $\delta\sum_{i=1}^{n-4}\delta = 4\epsilon$ and as $\delta,\epsilon \ra 0$ the probability of $e$ and $\pi = 124356\dots n$ will converge to the probabilities from the $n=4$ case, giving the same violation of $R$-decomposability when $\gamma_3\neq \gamma_4$. 
\endproof

Since both CDM and PCMC include all MNL models, we have the following corollaries.

\begin{corollary}
PCMC is not a reversible choice model for $n \ge 4$ items.
\end{corollary}

\begin{corollary}
CDM is not a reversible choice model for $n \ge 4$ items.
\end{corollary}

The fact that $RS$ and $RE$ produce different ranking distributions for non-reversible choice models and that popular choice models such as the MNL model are non-reversible can have very important consequences in practical modeling. For example, the widely-studied sushi dataset that consists of preference rankings over types of sushi was originally reported as having low ranks for items with low priority, but was later corrected so that low ranks represent high priority. Rankings built from these scores prior to the correction were thus ``backwards'' (see the warning by \citet{sushiwarning}). And as we've seen in this section, the difference between learning a ranking distribution from forwards vs.~backwards ranking data amounts to entirely different families of ranking distributions. Surprisingly, as part of our empirical results we find that the rankings provided in the sushi dataset (correctly ordered) are fit slightly better with $RE$ distributions than with $RS$ distributions.

\section{Markov chains on $S_n$}
\label{sec:mc-on-sn}
The study of self-organizing lists, originally motivated by research questions involving sequential access storage systems, introduced a number of Markov chains on $S_n$ that provide an altogether different way of defining distributions over rankings (through the stationary distributions of these chains). A self-organizing list is a storage model where a list reindexes its elements based on which indices are more commonly accessed, moving popular items towards the front to facilitate faster access. A popular method for managing a self-organizing list is the {\it move to front} (MTF) method, which simply places the most recently accessed item at the front of the list, sliding any item which had been ranked ahead of it back one position \citep{rivest1976self}. 

If we assume that each item $i$ is accessed with an independent probability $\gamma_i$, the MTF method induces a discrete time Markov chain on $S_n$ (as opposed to a chain on the set $U$, where the PCMC chain and other discrete choice chains live) where the stationary distribution of that chain describes the probabilities that the state of the list follows a given ranking of the items. In the language of choice modeling, the distribution of this chain is equivalent to a Plackett-Luce distribution with quality parameters $\gamma$. We now give a more complete explanation of this connection.

We define the MTF Markov chain as follows. For ranking $\sigma$ and alternative $i$ let $mtf(i,\sigma)$ be a mapping that returns a ranking $\sigma'$ with $\sigma'(i)=1, \sigma'(j)=\sigma(j)+1$ for $j$ such that $\sigma(j)<\sigma(i)$ and $\sigma'(j)=\sigma(j)$ for $j$ such that $\sigma(j)>\sigma(i)$. The transition probabilities of the Markov chain are thus
\begin{equation}
\label{eq:MTF}
P(\sigma,\sigma') = \gamma_i \1(\sigma' = mtf(i,\sigma)).
\end{equation}

We note that the Placett-Luce distribution, equivalent to repeated selection (RS) applied to the MNL model, can be written as a move to front chain.

\begin{proposition}[MTF yields Plackett-Luce distributions \citep{evans2014leading}]
\label{prop:MTF=PL}
The stationary distribution of an MTF chain with parameters $\gamma$ for items moving to the front is the same as the Plackett-Luce distribution with parameter vector $\gamma$:
\begin{equation}
\Pr(\sigma;\gamma) = \prod_{i=1}^n \frac{\gamma_{\sigma^{-1}(i)}}{\sum_{k\geq i} \gamma_{\sigma^{-1}(k)}}. 
\end{equation}
\end{proposition}

A notable extension of the MTF chain, for our interests, is the Markov Move to Front (MMTF) model \citep{dobrow1995move}, which has the same walk behavior as the MTF chain except the rate at which items move to the front of the chain depends on the current first item of the list (i.e.~the previous most recent item chosen). Thus instead of $n$ parameters $\gamma_1,...,\gamma_n$ there are $n^2$ parameters $\{a_{ij}\}_{i,j \in U}$ where $a_{ij}$ is the probability that $i$ is moved to the front of a ranking with $j$ currently in the front. The MMTF chain has the transition probabilities
\begin{equation}
\label{eq:MMTF}
P(\sigma,\sigma') = a_{i\sigma^{-1}(1)} \1(\sigma' = mtf(i,\sigma)).
\end{equation}

Although this distribution seems intuitively similar to RS applied to PCMC, where choices are made according to a random walk that has jump probabilities dependent on the current state of the PCMC chain, and for which Plackett-Luce is also a special case, the stationary distribution of an MMTF chain is generally {\it not} that of a repeatedly selected PCMC model.

\begin{proposition}[MMTF is RS only when it is MTF]
When there are at least four alternatives, the Markov Move-To-Front (MMTF) chain's stationary distribution is $L$-decomposable if and only if it is an MTF chain and has a Plackett-Luce distribution.
\end{proposition}

\proof
Let $\{a_{ij}\}$ be the parameters of an MMTF model, so that the probability that $i$ jumps to the front when $j$ is currently in the front is $a_{ij}$ as in Equation~\eqref{eq:MMTF}. 
We divide MMTF chains into two cases, one where $a_{ij}$ is fixed for all $j$, and another where it differs. 

If $a_{ij} = \gamma_i$ for every $j$ then the transition probabilities in Equation~\eqref{eq:MMTF} collapse to those of an MTF chain, as in Equation~\eqref{eq:MTF}. Then by Proposition~\ref{prop:MTF=PL}, it is also equal to a PL model and in turn a $RS_{PCMC}$ model. 

Now suppose that the $a_{ij}$ are not fixed for all $i$ for every $j$. The second item in an MMTF ranking is the item that was accessed before the first item. Let $\sigma$ be drawn from the stationary distribution of the MMTF. Then $\sigma^{-1}(1)$ is given by the state most recently visited by the chain on $U$. $\sigma^{-1}(2)$ is the second most recently visited state given $\sigma^{-1}(1)$ was the most recently visited. Note, however, that the chain may have self loops, but $\sigma^{-1}(1) \neq \sigma^{-1}(2)$, so more precisely, $\sigma^{-1}(2)$ was the state most recently visited before $\sigma^{-1}(1)$, not including $\sigma^{-1}(1)$. It further follows that $\sigma^{-1}(k)$ was the $k$-th most recently visited  {\it unique} state from a random walk on $U$. 

The distribution of $\sigma^{-1}(k)$ have been studied as {\it taboo probabilities} for the time-reversal of the chain on $U$, where the taboo probability of a state $i$ in a taboo set $S$ from a starting state $y \not \in S$ is the probability that a random walk hits $i$ before any other state in $S$ \citep{serfozo2009basics}.  Concretely, $Pr(\sigma^{-1}(k) = x | \sigma^{-1}(1),\dots,\sigma^{-1}(k-1))$ is the taboo probability of $x$ when the taboo set is $U - \{\sigma^{-1}(j)\}_{j\leq k}$ for a random walk starting in $\sigma^{-1}(k)$. 

Suppose we have at least four items, $w,x,y$ and $z$, and wlog $a_{wx} \neq a_{wy}$. Then when $\sigma(x)=1,\sigma(y)=2$, the taboo probability of $w$ for taboo set $U-\{x,y\}$ when starting from $x$ is different than when starting from $y$.
\endproof

As an informal corollary, repeated selection with the PCMC model and the Markov Move-To-Front chain produce the same ranking distribution if and only if both are some Plackett-Luce distribution.

While the MMTF chain does not fall within the ``choosing to rank'' framework, large-scale inference of these list-based Markov chains on $S_n$ is an interesting research direction. Further exploration of these processes, including the stationary distribution of MMTF chains, appear in \citet{bapat2003perron}. A major difficulty the prevents the practical learning of distributions for complex Markov chains on $S_n$ from ranking data is that many of these Markov chains have poor or unknown mixing times \citep{bhakta2013mixing}. 

\section{Mallows concentration parameter estimates}
\label{sec:mallows-mle}
Here we provide closed form estimates of the Mallow's concentration parameter for datasets with either full rankings or partial rankings. These estimates are derived given an estimate $\hat \sigma^*$ of the reference permutation, which may not be the maximum likelihood reference permutation (recall that finding the maximum likelihood permutation is NP-hard). Given an estimate $\hat \sigma^*$ we can find the maximum likelihood estimate of the concentration parameter {\it conditional} on $\hat \sigma^*$.

Recall that the Mallows model with concentration parameter $\theta$ and reference permutation $\sigma^*$ assigns each permutation $\sigma$ probability proportional to $\exp(-\theta \tau(\sigma,\sigma^*))$ where $\tau$ counts the number of inversions between $\sigma$ and $\sigma^*$, i.e.
$$
\tau(\sigma,\sigma^*) = \sum_{i<j} \1 \big [ \1 [\sigma(i)<\sigma(j) ] \ne \1 [\sigma^*(i)<\sigma^*(j) ] \big ],
$$
where $\1[\cdot]$ is the indicator function. 

We first consider full rankings. Given the reference permutation, the number of inversions in a sample from the Mallows model is thus binomially distributed Bin(${n \choose 2},p$), one trial for each pair and success probability $p=e^{-\theta}$ for each pair. Further, the total number of inversions over $k$ samples $\sigma_1,\dots,\sigma_k$ is binomially distributed as Bin($k{n \choose 2},e^{-\theta}$). It follows that the maximum likelihood estimate for $\theta$ can be derived from the corresponding Binomial distribution:
\[
\hat{\theta}^{MLE} = \log\left(\frac{\sum_{j=1}^k \tau(\sigma_k,\hat \sigma^*)}{k {n \choose 2}}\right).
\]

When our data includes partial rankings, we can still enumerate the number of pairs for which a partial ranking makes assertions, and compute the probability under which a partial ranking orients a pair differently than $\hat \sigma^*$. Assume $\sigma_j$ is of length $\ell_j$. Then $\sigma_j$ includes comparisons for ${\ell_j \choose 2} + (n-\ell_j)\ell_j$ pairs of items and then
\[
\hat \theta^{MLE}_{\text{partial}} = \log\left(\frac{\sum_{j=1}^k \tau(\sigma_k,\hat \sigma^*)}{\sum_{j=1}^k {\ell_j \choose 2} + (n-\ell_j)\ell_j}\right).
\]
For partial rankings we restrict the inversion count $\tau(\cdot,\cdot)$ to sum over those pairs for which the comparisons can be made.

\end{appendix}

\end{document}